\documentclass{article}

\usepackage[nonatbib, preprint]{neurips_2026}

\usepackage[utf8]{inputenc} 
\usepackage[T1]{fontenc}    
\usepackage{hyperref}       
\usepackage{url}            
\usepackage{booktabs}       
\usepackage{amsfonts}       
\usepackage{nicefrac}       
\usepackage{microtype}      
\usepackage{xcolor}         

\usepackage{hyperref}
\usepackage{color,soul}
\usepackage{graphicx}
\usepackage{tcolorbox} 
\usepackage{mdframed}
\usepackage{caption}
\usepackage{comment}
\usepackage{textcomp}
\usepackage[ruled,vlined]{algorithm2e}
\SetAlCapSkip{0.5em}
\SetAlgoInsideSkip{smallskip}
\SetAlgoHangIndent{0pt}
\SetKwInput{Problem}{Problem}
\SetKwInput{Parameter}{Parameter}
\SetKw{Initialize}{initialize}
\SetKw{Compute}{compute}
\SetKwInput{Extract}{Extract}
\usepackage{amsthm}
\usepackage{gensymb}
\usepackage{amsmath}
\usepackage{amssymb}
\usepackage{booktabs}
\usepackage{rotating}
\usepackage{enumitem}
\usepackage{nomencl}
\usepackage{adjustbox}
\usepackage{multirow}
\usepackage{float}
\usepackage{fontawesome5}
\usepackage{setspace}
\usepackage{mathtools, nccmath}
\usepackage{fancyhdr,lipsum}
\usepackage{subcaption}
\usepackage{placeins}
\usepackage{textcomp}

\usepackage{subcaption}
\usepackage{titlesec}
\usepackage{textcomp}
\usepackage{lscape}
\usepackage{wrapfig,lipsum,booktabs}
\usepackage{booktabs, ltablex, makecell}
\usepackage[none]{hyphenat}
\usepackage{eucal}
\usepackage{lineno}
\title{\texttt{NLPOpt-Net}: A Learning Method for Nonlinear Optimization with Feasibility Guarantees}

\author{%
Bimol Nath Roy \\
Texas A\&M University\\
\texttt{bimolnathroy@tamu.edu}
\And
Rahul Golder \\
Texas A\&M University\\
\texttt{rahulgolder8420@tamu.edu}
\And
M. M. Faruque Hasan\thanks{Corresponding author} \\
Texas A\&M University\\
\texttt{hasan@tamu.edu}
}

\begin{document}

\maketitle

\begin{abstract}
Nonlinear Parametric Optimization Network (\texttt{NLPOpt-Net}) is an unsupervised learning architecture to solve constrained nonlinear programs (NLP). Given the structure of an NLP, it learns the parametric solution maps with guaranteed constraint satisfaction. The architecture consists of a backbone neural network (NN) followed by a multilayer ($k$-layered) projection. While the NN drives toward optimality through a loss function consisting of a modified Lagrangian augmented with a consistency loss, the projection ensures feasibility by projecting the NN predictions in the original constraint manifold. Instead of typical distance minimization, our projection exploits local quadratic approximations of the original NLP. Under certain conditions (such as convexity), the projection has a descent property, which improves the NN predictions further. \texttt{NLPOpt-Net} deploys an inversion-free, modified Chambolle-Pock algorithm to solve the constrained quadratic projections during the forward pass and uses the implicit function theorem for efficient backpropagation. The fixed structure of the projection further allows decoupling of the NN and the projection once the training is complete. \texttt{NLPOpt-Net} solves large-scale convex QP, QCQP, NLP, and nonconvex problems with near zero optimality gap and constraint violations reduced to machine precision. Additionally, it provides near accurate prediction of the active sets and corresponding dual variables, thereby enabling a scalable approach for multiparametric programming. Compiling the projection in C provides order of magnitude improvement in inference time compared to JAX. We provide the codes and \texttt{NLPOpt-Net} as a ready to use package that includes GPU support.
\end{abstract}
\faGithub\,\,\, \href{https://github.com/souls-tamu/nlpoptnet}{\texttt{https://github.com/souls-tamu/nlpoptnet}} \hfill \faPython\,\,\, \href{https://pypi.org/project/nlpoptnet/}{\texttt{pip install nlpoptnet}}
\section{Introduction} \label{intro}
We consider the following nonlinear optimization (NLP) problem:
\begin{ceqn}
    \begin{equation}\label{Problem} \tag{P}
        \begin{aligned}
            \min f({x,y}),\quad \mathrm{s.t.}\ {h}({x},{y}) = {0}, \quad {g}({x,y})\le {0}, \quad {l}({x}) \leq {y} \leq {u}({x}),
        \end{aligned}
    \end{equation}
\end{ceqn}
where, $y$ are the decision variables and $x$ are the parameters. We aim to find the optimal solutions of P for different realizations of $x$. Formally, our goal is to train a neural network (NN), ${\hat{y}}=\Phi_{NN}({x};{\Theta})$, and then project $\hat{y}$ onto the constraint set such that projected output $\tilde{y}$ always belong to the feasible set $\mathcal{S}$, i.e., $\tilde{y}\in\mathcal{S}$, where $\mathcal{S}=\{{y}\ \vert\ {h}({x},{y}) = {0},\ {g}({x,y})\le {0},\ {l}({x}) \leq {y} \leq {u}({x})\}$. Here, ${h}$ represents a set of parameterized equality constraints $h_i({x},{y})=0,\ \forall i\in\mathcal{N}_E$; ${g}$ represents a set of parameterized inequality constraints $g_j({x,{y}})=0,\ \forall j\in\mathcal{N}_I$; and ${l}$ and ${u}$ represent the parameterized lower and upper bounds of the variables ${y}$, respectively. Lastly, ${\Theta}$ represents the aggregated tunable parameters of the NN.

NLPs are used to optimize  systems governed by nonlinear objectives and/or constraints that arise in a broad range of applications such as robotics and control, energy systems optimization, operations research,  advanced manufacturing, materials science, and many others. There are well-established techniques (see \cite{nocedal2006numerical}) and packages (e.g., IPOPT \cite{wachter2006implementation}, CONOPT \cite{drud1985conopt}, Knitro \cite{byrd2006knitro}, SNOPT \cite{gill2005snopt}) for solving NLPs. 
However, they are primarily designed to solve a single NLP instance robustly, but are not well-suited for repeatedly solving the same NLP many times in applications such as real-time optimization, model predictive control, and design and optimization under uncertainty. 
One way to address this challenge is by deriving the solution maps \textit{a priori} across varying parameters through parametric optimization \cite{bank1982non,pistikopoulos2020multi}. 
While explicit parameter-to-solution mappings exist for certain problem classes, such as linear and quadratic programs \cite{bemporad2002explicit, kenefake2023multi}, conventional approaches for repeatedly solving constrained NLPs are computationally intensive and lack scalability.
Deep NNs can be trained offline over a distribution of parameters or problem instances to ``learn'' the solution maps of a parametric optimization problem. However, NNs inherently lack mechanisms to ensure explicit constraint satisfaction, a limitation that motivates a large body of existing work.


Penalizing constraint violations through the NN loss function is a simple and practical approach often applied in parametric optimization \cite{drgona2023neuromancer} and physics informed neural networks (PINNs) \cite{raissi2019physics, marquez2017imposing, erichson2019physics}. Despite being flexible and generalizable, soft-constrained methods suffer from an inherent tradeoff between prediction accuracy and constraint feasibility \cite{cai2021physics}. Manual tuning is required to adjust the weights of the penalty terms, which can alter the solution \cite{grontas2026operator}. Even with high penalties, constraint violation can still persist \cite{wang2022and, ma2022data}. 
In parametric optimization, when constraints are treated softly, the prediction may admit solutions that are not physically realizable, especially for problems arising in systems that are well defined by physical laws.
Hard-constrained NNs focus on methodological approaches that can ensure constraint feasibility within a tolerance. These can be broadly classified into two categories. The first is \textit{predict-and-complete} method \cite{beucler2021enforcing, zamzam2020learning} where a subset of the outputs is predicted through the NN and the rest is computed by solving the system of equality constraints. 
Donti et al. \cite{donti2021dc3} in their work \texttt{DC3}, introduced a framework that uses predict-and-complete method for equality constraints and a correction strategy to satisfy the inequality constraints. 
The second method is \textit{projection} that incorporates a differentiable projection layer to project the outputs of an NN into the feasible constraint manifold. Early work such as OptNet \cite{amos2017optnet} introduced quadratic programs as differentiable layers, later extended to broader classes of disciplined convex programs \cite{agrawal2019differentiable}. Chen et al. \cite{chen2024physics}  proposed a Karush-Kuhn-Tucker (KKT) condition-based orthogonal projection with explicit analytical form for enforcing linear constraints. Iftakher et al. \cite{iftakher2025physics} developed \texttt{KKT-HardNet}, a general framework that finds a stationary point for the KKT system of projection problem to satisfy both nonlinear equality and inequality constraints. This approach is later extended to satisfy differential-algebraic (DAE) constraints in \texttt{DAE-HardNet} \cite{golder2025dae}. Grontas et al. \cite{grontas2025pinet} introduced \texttt{Pinet}, an operator splitting based projection method that can incorporate convex constraints. Consante et al. \cite{constante2025enforcing} proposed decision rule-based networks to incorporating linear inequality constraints. Lastrucci et al. \cite{lastrucci2025enforce} used adaptive neural projection (adaNP) that recursively linearizes nonlinear equality constraints. A common trait of the projection-based method is that the projection is based on Euclidean distance minimization. For learning the solutions of constrained optimization problems, this approach tends to move away from the true optima. To the best of our knowledge, no existing method addresses this.

In this work, we introduce Nonlinear Parametric Optimization Network (\texttt{NLPOpt-Net}), a general deep learning tool for solving convex NLPs. The architecture consists of a backbone NN followed by a multilayer ($k$-layered) projection. While the NN drives toward optimality through a loss function consisting of a modified Lagrangian augmented with a consistency loss, the projection ensures feasibility in the original constraint manifold. Specifically, we linearize the constraints at NN output and instead of traditional distance minimization projection, we quadratically approximate the objective such that the quadratic coefficient remains diagonal. This approximated QP is then solved with an inversion-free method. Instead of moving away the optima, this approximation in the projection guides the NN to find better optima. For affine constraints, a single approximation is sufficient. For general nonlinear constraints, we repeat this process for $k$ times and we provide convergence guarantees after $k$ layers of projection to attain a feasible NLP solution. While, backbone NN improves initialization for faster convergence, the projection has a descent property which allows to find a no-worse objective after projection. This descent property further pushes the framework towards optimality. The $k$-layered projection can be incorporated with any standard deep learning architectures. Moreover, the projection has a computationally cheap forward pass and the backward propagation is expedited with gradient computation through implicit function theorem. \\

\begin{mdframed}[
    backgroundcolor=blue!3,
    linecolor=blue!5,
    linewidth=0pt,
    skipabove=0pt,
    skipbelow=0pt
]
Our key methodological contributions include: 
\textbf{(1)} a general deep learning framework, NLPOpt-Net, for learning the  solutions and the active sets of parametric nonlinear optimization problems with a $k$-layered projection that ensures constraint feasibility; and \textbf{(2)} an first order primal-dual algorithm for solving the projection problems with high efficiency and a descent property when the problem is convex; \textbf{(3)} computational studies show success over other methods for different convex NLP and noncovex cases with superior results in terms of optimality and strict constraint satisfaction; and \textbf{(4)} instead of relying on just-in-time (JIT) compilation for the entire architecture, we decouple the projection from the NN and utilize the ahead-of-time (AOT) compilation for the projection to accelerate the single instance inference for practical deployment. We also provide a ready to use GPU supported package.
\end{mdframed}

\section{NLPOpt-Net: Theoretical Developments}\label{ThDv}

\begin{figure}[h]
    \centering
    \includegraphics[width=0.70\linewidth]{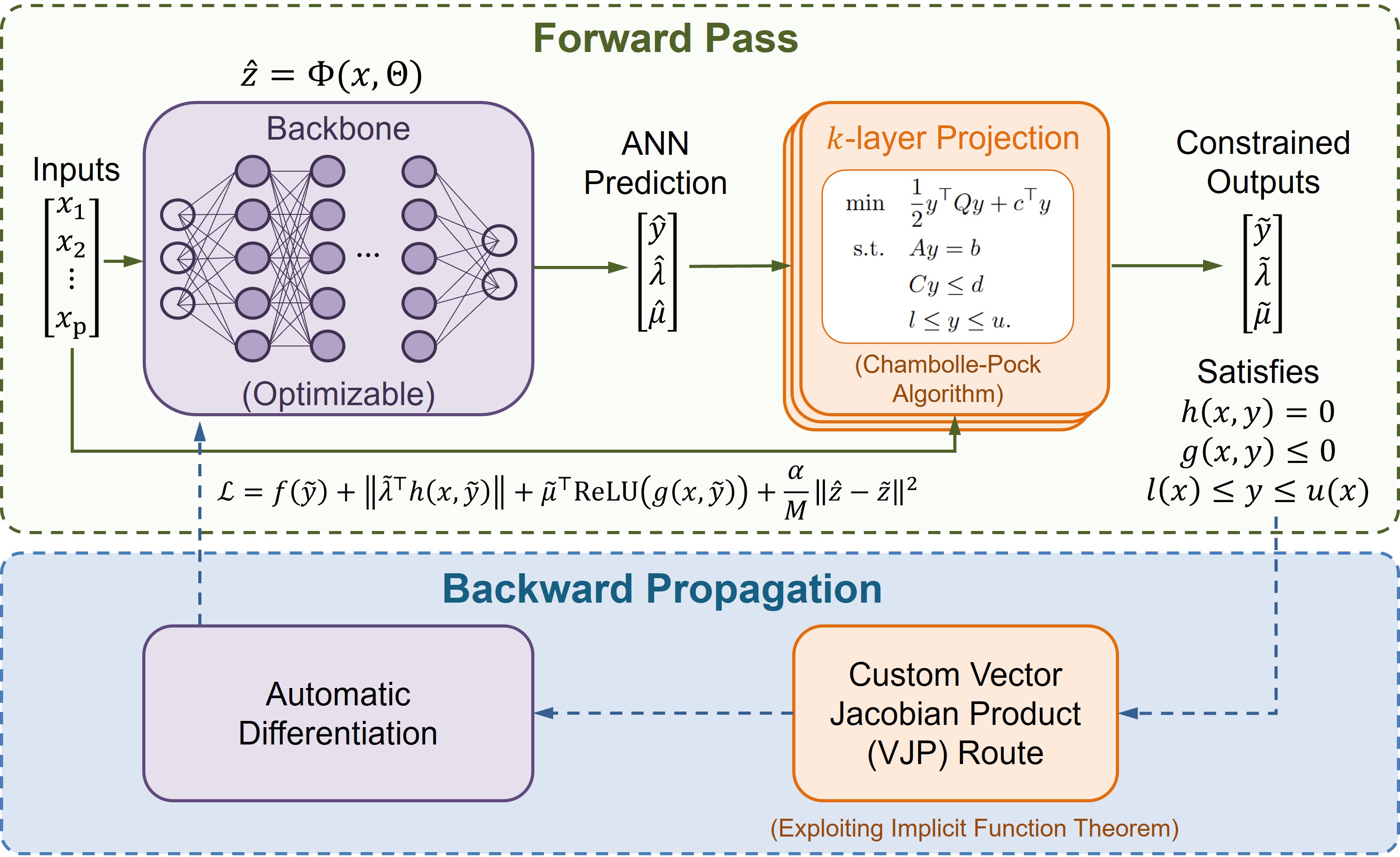}
    \caption{Illustration of the \texttt{NLPOpt-Net} framework. The forward pass includes NN followed by $k$-layered projection. The backward pass uses custom VJP route. The NN is optimized with the original objective along with a regularized soft loss and a consistency loss.}
    \label{fig:nlpoptnet}
\end{figure}

The \texttt{NLPOpt-Net} (Nonlinear Parametric Optimization Network) framework is illustrated in Fig. \ref{fig:nlpoptnet}. The architecture of NLPOpt-Net includes (a) a backbone NN, which can be of any kind, (b) a differentiable projection layer which can have multiple sub-layers or single sub-layers (depending on the nonlinearity). We utilize automatic differentiation for backward propagation of the backbone and custom defined backward gradient for the projection layer. In the forward pass the backbone uses standard computation and the projection is solved with a Modification of the Chambolle-Pock (CP) algorithm\cite{chambolle2011first}. During forward pass of subproblem, the Jacobian is computed with automatic differentiation which is computationally cheap. Furthermore, all the computations can be parallelized in GPU. Our framework is able to train directly from the problem specification, i.e., the framework is fully unsupervised. Furthermore, when the NN prediction becomes close, the loss becomes self-supervised due to the consistency loss. We incorporate a first order primal dual method, the Chambolle-Pock algorithm to run the projection layer. We employ the projection for $k$-layers to converge to a feasible solution. Moreover, if the constraints are affine the projection become minimal.
\subsection{Backbone Neural Network} 
We consider a standard artificial neural network (ANN) with ReLU activation function. Considering, $z = [y^\top,\lambda^\top,\mu^\top]^\top$, a vector defining all the primal and dual variables, the loss function $\mathcal{L}$, a modified lagrangian augmented with a consistency term, has the following general form:
\begin{ceqn}
    \begin{equation}
        \mathcal{L} = f(\tilde{y}) + \lVert \tilde{\lambda}^\top h(x,\tilde{y}) \rVert^2 + \tilde{\mu}^\top \mathrm{ReLU}\big(g(x,\tilde{y})\big) + \frac{\alpha}{M}\lVert \hat{z}-\tilde{z}\rVert^2,
    \end{equation}
\end{ceqn}
The only user defined tuning parameter here is $\alpha$ which enforces the consistency loss. The consistency loss considers the minor linearization error and keep the unconstrained prediction in the local neighborhood of the original solution. The loss function pushes the NN prediction to get close to the optimal solution of (\ref{Problem}). However, due to the soft embedding of the constraints through penalties, we do not have either guarantees of feasibility and optimality. For this reason, we next opt for projection which ensures feasibility with a descent property. As we go through training, the backbone NN with the projection layers is expected to get close to the actual solution of (\ref{Problem}).
\subsection{Projection}\label{NLPOpt}
The projection-based constraint satisfaction can be done by projecting the backbone prediction, $\hat{{y}}$ onto the feasible constraint manifold $\mathcal{S}$ \cite{chen2024physics, lastrucci2025enforce, iftakher2025physics, grontas2025pinet}. While projecting with distance minimization ensures feasibility, it may move $\tilde{y}$ from the optimal solution $y^*$. Instead of solving a distance minimization problem in the projection layer, we solve a local quadratic approximation of the objective subject to the constraints as shown in \ref{Projection}. We show that the approximation at the optimal point has the same primal-dual solution as the original problem \ref{Problem}.
\begin{ceqn}
    \begin{equation}\label{Projection} \tag{PR}
            \tilde{{y}} = \arg \min_{{y}\in\mathcal{S}} \quad f({x},\hat{{y}}) + J^\top({y}-\hat{{y}}) + \frac{1}{2}({y}-\hat{{y}})^\top H_{d}({y}-\hat{{y}}),
    \end{equation}
\end{ceqn}
where, $J = \nabla f({x}, \hat{{y}})$, $H_{d}=\rho \ \mathrm{diag}\left(\nabla^2 f({x},\hat{{y}})\right)$ and $\rho$ is a parameter (in most cases $\rho=1$, we discuss when $\rho$ differs to preserve descent property). Here, we notice the similarity to a proximal second order approximation of the objective used in augmented Lagrangian methods\cite{neal2011distributed}. However, unlike proximal approximation, we rely on different weight multipliers for different variables which are Hessian dependent. We make the following assumptions: (i) The number of equality constraints, $\lvert\mathcal{N}_E\rvert\le n$, (ii) Each equality constraint, ${h}$ is affine in $y$ and linearly independent and the constraint ${g}$ is convex nonlinear in ${y}$. ${h},\ {g}$ are continuously differentiable, (iii) $f$ is twice continuously differentiable and convex and its gradient is Lipschitz continuous with Lipschitz constant $L$ and (iv) All the optimization instances are feasible for any feasible realization of $x$. The projection problem of $\min_{y\in\mathcal{S}} f(x,y)$ is $\min_{y\in\mathcal{S}} \hat{f}(x,y)$ where $\hat{f}(x,y) = f({x},\hat{{y}}) + J^\top({y}-\hat{{y}}) + \tfrac{1}{2}({y}-\hat{{y}})^\top H_{d}({y}-\hat{{y}})$. As per assumption of $L$-continuous gradient,
\begin{ceqn}
    \begin{equation*}
        f(x,y) \le f(x, \hat{y}) + \nabla f(x,\hat{y})^\top (y-\hat{y}) + \frac{L}{2}\lVert y-\hat{y}\rVert^2.
    \end{equation*}
\end{ceqn}
For $H_d\succeq LI$, $({L}/{2})\lVert y-\hat{y}\rVert^2\leq (1/2) (y-\hat{y})^\top H_d(y-\hat{y})$. Therefore, $f(x,y)\leq \hat{f}(x,y)\ \forall y$. Since $\tilde{y}$ minimizes $\hat{f}(x,y)$ over $\mathcal{S}$, we have $\hat{f}(x,\tilde{y})\leq \hat{f}(x,\hat{y})$ i.e., $\hat{f}(x,\tilde{y})\leq {f}(x,\hat{y})$ ($f$ and $\hat{f}$ are exact at $\hat{y}$). This yields, $f(\tilde{y})\leq f(\hat{y})$. This infers that the projection \ref{Projection} yields a descent point $\tilde{y}$ when $H_d\succeq LI$, i.e., $f(\tilde{y})\leq f(\hat{y})$. It is important to note that the condition $H_d\succeq LI$ is a weak sufficient condition for descent property but not a necessary condition. We consider $H_d=\rho\ \mathrm{diag}\left(\nabla^2f(x,\hat{y})\right)$. For most of the cases $\rho=1$ retains the descent property, especially when the hessian is diagonally dominant. However, a weak bound on $\rho$ can be imposed with $\rho\geq\frac{L}{\min_i\partial^2_{ii}f(x,\hat{y})}$. While automation can be done by detecting non-descent evaluation before and after projection, in our implementation we use a constant $\rho$ provided by user.

\textbf{$k$-Layered Projection:}
A local solution of the nonlinear problem \ref{Projection} can be obtained by solving the KKT conditions. This is apparently the global solution as the problem remains convex under the previous assumptions. However, finding the KKT solution is not straightforward and involves solving a system of nonlinear equations (more details can be found in \cite{nocedal2006numerical}). To that end, we locally approximate the nonlinear projection problem (\ref{Projection}) using the following (\ref{subproblem}):
\begin{ceqn}
    \begin{equation}\label{subproblem} \tag{SP}
        \begin{aligned}
            \min &\quad\frac{1}{2}{y}^\top Q{y} + c^\top {y},\quad\mathrm{s.t.}\ A{y} = b,\ C{y} \leq d, \ l \leq {y} \leq u.
        \end{aligned}
    \end{equation}
\end{ceqn}
Here, the objective of \ref{subproblem} is second order approximation of the original objective with only diagonal Hessian elements, and the constraints are first order approximation at  point $({x},{y}^i)$, where $i = 1, ..., k$ in a $k$-layered projection. Mathematically, $Q = \rho\,\mathrm{diag}(\nabla^2 f({x},{y}^i))$, $c = \nabla f({x},{y}^i) - \rho\,\mathrm{diag}(\nabla^2 f({x},{y}^i)){y}^i$, $A = J_{{h}}({x},{y}^i)$, $b = J_{{h}}({x},{y}^i){y}^i - {h}({x},{y}^i)$, $C = J_{{g}}({x},{y}^i)$, $d = J_{{g}}({x},{y}^i){y}^k - {g}({x},{y}^i)$, $l = {l}({x})$ and $u = {u}({x})$, where, $\nabla^2f$ is the Hessian of the objective, diag$(\cdot)$ represents the matrix formed with diagonal element of a matrix, $\nabla f$ is the gradient of the objective, and $J_{{h}}$ and $J_{{g}}$ are the Jacobian of the equality and inequality constraints, respectively. Note that we start from the backbone prediction, i.e., ${y}^0=\hat{{y}}$. We sequentially solve \ref{subproblem}$_i$ approximated at $({x},{y}^i)$ in  each projection layer $i$ for $i = 1, ..., k$.

The multi-layer projection is used in a sequential manner to get to the feasible solution. We define an operator $\mathcal{P}_s$ such that $\tilde{{y}}=\mathcal{P}_s({{x}},\hat{{y}})$ is the solution to the approximated subproblem \ref{subproblem} at the point of approximation being $({{x}},\hat{{y}})$. Given, the inversion free iterative scheme of the operator $\mathcal{P}_s$ provided in section \ref{Chambolle-Pock}, we define an overall projection layer represented by $\mathcal{P}_o$. It is trivial to show that under assumptions made earlier, \ref{subproblem} remains feasible and has a primal-dual solution of \ref{subproblem}. It can be said that a single $\mathcal{P}_s$ cannot ensure feasibility in the original nonlinear constraint manifold. However, a linear convergence rate under certain assumption can be achieved for running sequential subproblems (proof in Appendix \ref{local_convergence}). When the backbone prediction is close to the original solution, the number of required sequential subproblems becomes minimal. The goal is to train the backbone to predict such neighborhood point so that with minimal change feasibility and optimality can be restored. We define the complete projection $\mathcal{P}_o$ as a composition of sequential operators $\mathcal{P}_s$ implemented in sequential layers, such that, $\mathcal{P}_o=\mathcal{P}_s^1\circ\mathcal{P}_{s}^2\circ\cdots\mathcal{P}_{s}^k$, where, $\mathcal{P}_o$ ($k$-layered Projection) represents solving \ref{Projection} as QP, and $\mathcal{P}_s^i$ represents solving the subproblem \ref{subproblem} in $i$th layer. Given an arbitrarily small scalar $\epsilon$ and $k\in\mathbb{N}$, $\tilde{{y}}$ is computed from the backbone prediction $\hat{{y}}$ as $\tilde{{y}} = \mathcal{P}_o(x, \hat{{y}})$, and converges to an optimal solution of \ref{Projection} with linear convergence rate under constraint smoothness conditions such that $\hat{{y}}\in\mathcal{S}$ with a maximum absolute violation of $\epsilon$ or less.
\subsection{Solving SP in the Projection Layer} \label{Chambolle-Pock}
While many methods exist for solving QPs, we utilize the efficiency of the Chambolle-Pock (CP) algorithm \cite{chambolle2011first} (see Algorithm \ref{CP_QP}) to solve the QP in each $\mathcal{P}_s^i$. The CP algorithm is a first-order primal dual algorithm that has resemblance to Douglas Rachford (DR) splitting algorithm and can be understood as a preconditioned version of the alternating direction method of multipliers (ADMM) \cite{chambolle2011first}. We provide the step by step updates, the termination criteria and convergence of the CP algorithm for the problem defined in Eq. \ref{QP_Saddle} in Appendix \ref{stepbystepCP}.

\begin{algorithm}[H]
\small
\caption{CP for QPs}\label{CP_QP}
\DontPrintSemicolon
\SetKwBlock{CPQP}{\texttt{ChambollePock}:}{}
\CPQP{
\KwIn{Initial values $y^0,\lambda^0,\mu^0 $, $p$}
\Problem{$\min \tfrac{1}{2}y^\top Qy+c^\top y;\quad \mathrm{s.t.}\ Ay=b; \ Cy \leq d; \ l\leq y\leq u$ }
\Extract{$Q, c, A, b, C, d, l ,u \leftarrow p$}
\Parameter{$\tau>0, \sigma>0, \theta=1$ while $\left(\frac{1}{\tau}-L_f\right)\frac{1}{\sigma}<\lVert K\rVert^2$}
\Compute{$P=\mathrm{diag}(1/(1+\tau q_i))$}\;
\Initialize{$k\leftarrow0, y^0\leftarrow y^0,\lambda^0\leftarrow \lambda^0,\mu^0\leftarrow \mu^0,\bar{y}^0\leftarrow y^0$}\;
\Repeat{termination criteria is satisfied}{
        $\lambda^{k+1} = \lambda^k + \sigma (A\bar{y}^k-b);\quad \mu^{k+1} = \max(0,\mu^k + \sigma (C\bar{y}^k-d))$\;
        $\nu^{k+1} = P\big(y^k - \tau (A^\top\lambda^{k+1}+C^\top\mu^{k+1})-\tau c\big);\quad y^{k+1} = \Pi_{[l,u]}(\nu^{k+1})$\;
        $\bar{y}^{k+1}=y^{k+1}+\theta(y^{k+1}-y^k)$\;
        $k\leftarrow k+1$
        }
\KwOut{$z\leftarrow y^{k},\lambda^k,\mu^k$}}
\end{algorithm}

\subsection{Backward Gradient Computation}
We predict both primal and dual variables and use them in the training loss. With $z = [y^\top, \lambda^\top,\mu^\top]^\top$, the loss function $\mathcal{L}(x,z)$, the projection layer $\mathcal{P}_o(x,z)$, and the NN $\Phi(x; \Theta)$, the loss function can be generalized into as a composite function of $\Theta$ and $x$ such that $\mathcal{L}(x,\tilde{z})=\mathcal{L}(x,\mathcal{P}_o(x,\Phi(x;\Theta)))$. To train the backbone NN with backpropagation, the loss $\mathcal{L}$ has to be differentiated with respect to backbone parameters $\Theta$. Using chain rule, we get
\begin{ceqn}
    \begin{equation}
    \frac{\mathrm{d}\mathcal{L}(x,\mathcal{P}_o(x,\Phi(x;\Theta)))}{\mathrm{d}\Theta}=\frac{\partial\mathcal{L}(x,z)}{\partial z}\Bigg|_{z=\tilde{z}} \ \frac{\partial \mathcal{P}_o(x,z)}{\partial z}\Bigg|_{z=\hat{z}} \ \frac{\partial \Phi(x;\mathcal{\vartheta})}{\partial \vartheta}\Bigg|_{\vartheta=\Theta},
    \end{equation}
\end{ceqn}
where, $\hat{z}=\Phi(x; \Theta)$ and $\tilde{z}=\mathcal{P}_o(x,\hat{z})$. The projection layer is solved in an iterative scheme. While the first and last terms are standard and efficiently computed with automatic differentiation, computing the backward gradient with automatic differentiation for the second term suffers due to the large number of iterations in the projection layer. We utilize the implicit function theorem to compute the backward gradient of the CP algorithm. The detailed derivation of the backward gradient for the projection is given in Appendix \ref{backwardGrad}. This gives us the leverage to use this method as a differentiable layer inside a NN framework with efficient training time. The use of the implicit function theorem in backward gradient is not new for iterative processes. For example, Optnet \cite{amos2017optnet} and Pinet \cite{grontas2025pinet} used implicit backward gradient for iterative schemes such as the Newton's method and the Douglas-Rachford method.

\subsection{Training NLPOpt-Net}\label{loss_function}
We train NLPOpt-Net with standard algorithms (i.e., Adam \cite{kingma2014adam}). The heuristic use of the dual variables in the loss function trains the model to lower the optimality gap. The training/testing and the overall NLPOpt-Net framework are given in Algorithm \ref{NLPOpt_framework} and \ref{trainingtesting}.
\begin{figure*}[h]
\centering
\begin{minipage}[t]{0.51\textwidth}
\raggedright
\begin{algorithm}[H]
\caption{NLPOpt}\label{NLPOpt_framework}
\DontPrintSemicolon
\SetKwBlock{fwd}{\texttt{Forward}:}{}
\fwd{
\KwIn{$x,\Theta,n_{\max}$}
$\hat{z}\leftarrow \Phi(x;\Theta),\ z^0\leftarrow \hat{z}$\; 
\For{$k=0$ \KwTo $n_{\max}-1$}{
    $p^k\leftarrow \mathcal{M}(x,z^k)$\;
    $(y^0,\lambda^0,\mu^0)\leftarrow z^k$\;
    $z^{k+1}\leftarrow \texttt{ChambollePock}(y^0,\lambda^0,\mu^0,p^k)$
}
$\tilde{z}\leftarrow z^k$\;
\KwOut{$\hat{z}, \tilde{z}$}}
\vspace{0.3em}
\SetKwBlock{bwd}{\texttt{Backward}:}{}
\bwd{
\KwIn{$\tilde{z},\ \hat{z},\ x,\ g_z=\dfrac{\partial \mathcal{L}}{\partial \tilde{z}}$}
$v \leftarrow \texttt{Solve}\!\left((I-J_F(z^*)^\top)v=g_z\right)$\\
$\xi \leftarrow v^\top \dfrac{\partial F}{\partial p}\Big|_{p=\mathcal{M}(x,\hat{z})}$\\
$w \leftarrow \xi \dfrac{\partial \mathcal{M}(x,z)}{\partial z}\Big|_{z=\hat{z}}$\\
$\dfrac{d\mathcal{L}}{d\Theta}
\leftarrow
w \dfrac{\partial \Phi(x;\Theta)}{\partial \Theta}$\\
\KwOut{$\dfrac{d\mathcal{L}}{d\Theta}$}}
\end{algorithm}
\end{minipage}
\hfill
\begin{minipage}[t]{0.46\textwidth}
\begin{algorithm}[H]
\small
\caption{Training/Testing}\label{trainingtesting}
\DontPrintSemicolon
\KwIn{Chosen/tuned hyperparameters}
\SetKwBlock{train}{\texttt{Training}:}{}
\train{
\Initialize{$\Theta_1$}\;
\For{$t=1$ \KwTo $n_{\mathrm{epochs}}$}{
    $x \leftarrow$ sample from training set\;
    $\hat{z},\tilde{z} \leftarrow \texttt{Forward}(x,\Theta_t,n_{\max})$\;
    \Compute{$\mathcal{L}(x,z)$}\;
    $\dfrac{\partial \mathcal{L}}{\partial \Theta_\ell}
    \leftarrow
    \texttt{Backward}\!\left(\tilde{z},\hat{z},x,\dfrac{\partial \mathcal{L}}{\partial \tilde{z}}\right)$\;
    $\Theta_{\ell+1}\leftarrow \mathrm{Update}(\texttt{adam})$\;
}}
\SetKwBlock{test}{\texttt{Testing}:}{}
\test{
$x \leftarrow$ sample from test set\;
$\hat{z}, \tilde{z}\leftarrow \texttt{Forward}(x,\Theta_{n_{\mathrm{epochs}}},n_{\max})$\;}
\KwOut{$\Theta_{n_{\mathrm{epochs}}},\ \tilde{z}$}
\end{algorithm}
\end{minipage}
\end{figure*}

\subsection{Package and Implementation}\label{package}
We provide a GPU ready user package with pip installation. The projection can be further optimized once the training is complete. In practical setting, single inference is more common than batch inference. With that concern, we provide two options for the projection step during inference. The default (\texttt{`jax'}) which is implemented in JAX. After training is finished, a native C compiled projection (\texttt{`native'}) can alternatively accelerate single inference. This is achieved by decoupling the NN and projection computations. The NN is executed using JIT compilation while the projection is deployed via AOT compilation. We find that in practical setup a trained model single inference for a QP can be done with the same order of magnitude time as state of the art optimizer (e.g., \texttt{OSQP}). \footnote{The result in Table \ref{tab:package_result} can be found in the notebook(\texttt{Inference\_comparison.ipynb}) in GitHub repository.} Note that, this experimentation was carried out on a 13\textsuperscript{th}~Gen Intel\textsuperscript{\textregistered} Core\textsuperscript{\texttrademark} i7--13700 (24-core CPU) with 32 GB RAM running Linux OS for reflecting realistic deployment scenarios. Table \ref{tab:package_result} provides the average time of inference and average optimality gap over all 2000 instances. We find that the native inference method achieves $\approx50\times$ speedup than the JAX method. Note that the time for OSQP is only the solve time which does not include the problem creation overhead where as the time for NLPOpt is the end-to-end inference time after training.
\begin{table}[h]
    \caption{Overall summary of the QP used to compare the single instance with the implemented package with state-of-the-art solver OSQP. The metrics are the inference time for a single instance and the optimality gap. Both metrics are provided as an average over the 2000 individual instances including the training and validation.\\}
    \label{tab:package_result}
    \centering
    \begin{tabular*}{\linewidth}{@{\extracolsep{\fill}}lccc}
    \toprule
    Metric  & Optimizer (\texttt{OSQP}) & NLPOpt-Net(\texttt{`native'}) & NLPOpt-Net(\texttt{`jax'}) \\ \midrule
    Avg. inf. time/instance & 2.22 ms   & 2.06 ms       & 95.14 ms       \\
    Average optimality gap           & 0.000\%   & 0.081\%       & 0.081\%    \\ \bottomrule
    \end{tabular*}
\end{table}
\section{Computational Studies}\label{ComStd}
We deploy NLPOpt-Net for learning to solve convex Quadratic Programs (QP), convex Quadratically Constrained Quadratic Programs (QCQPs) and a family of convex nonlinear programs (NLPs). Despite the convergence and theoretical descent being established for convex cases, we also test the framework on non-convex NLPs.\footnote{Codes for all experiments are available at \texttt{\hyperlink{https://github.com/souls-tamu/nlpoptnet}{\texttt{https://github.com/souls-tamu/nlpoptnet}}}.} All the experiments carried out in this section have 100 variables, and 50 equality and 50 inequality
constraints and 50 parameters. After training, we load the models and test on all the instances. We compare the performances on the following metrics:

\textbf{Optimality:} Optimality refers to how close the final prediction is to the true optimal solution. We measure the average optimality gap (AOG) as the following formula:
\begin{ceqn}
    \begin{equation*}
        \textrm{AOG} = \frac{\textrm{Average Objective Value (\textit{Model Objective})}-\textrm{Best Bound (\textit{Optimizer Objective})}}{\lvert \textrm{Average Objective Value} \rvert}\times 100.
    \end{equation*}
\end{ceqn}

\textbf{Feasibility:} Feasibility is measured by the constraint violation. The constraint violation is computed as the absolute residual for equality constraint, i.e, $\lvert h(x,y)\rvert$ and $g_+(x,y)$ for inequality where $g_+ = \max(0,g)$. We evaluate the maximum violation and mean violation across all the instances.

\textbf{Speed:} Speed refers to the time required for inference after the training is finished. The inference time is given for each batch. For optimizer we consider full parallelization and for a fare comparison we divide the total solve time by total number of instances in a batch (We consider a batch size of $400$ instances for all experiments).

We compare NLPOpt-Net against the following methods:

\textit{\textbf{Optimizer:}} A traditional optimization solver. We use CVXPY\cite{cvxpy} environment (for convex problem, solver: \texttt{OSQP}\cite{stellato2020osqp} for QPs, \texttt{SCS}\cite{scs} for QCQPs and NLPs) and \texttt{SLSQP}\cite{2020SciPy-NMeth} (the SciPy implementation using \texttt{scipy.optimize} for nonconvex problems).

\textit{\textbf{NN:}} A regular ANN model with soft loss for constraint violation. 

\textit{\textbf{Eq. NN:}} This supervised learning \cite{zamzam2020learning} predicts and completes the full set of variables with equality constraints. The model is trained on minimizing the mean squared error between the optimal solutions of the Optimizer and the model completed solution.

\textit{\textbf{DC3:}} The deep constraint completion and correction (DC3) \cite{donti2021dc3} method is state-of-the-art that uses predict and completion method to satisfy the equality constraints. The inequality correction is then done by mapping infeasible points to feasible points using an internal gradient descent step during training. We use the github codespace for DC3 including default tunable parameters for comparison.

All experiments are executed on a computing system featuring 48 CPU cores from dual Intel Xeon 6248R 3.0GHz processors, 1 NVIDIA A100 40 GB GPU node, and 360 GB RAM allocated per run. For cases that are run sequentially in CPU, we divide the total time for each batch by the total number of problem instances in a batch considering full parallelization. We run each experiment 5 times for 2000 epochs with  different seeds (with a train/test ratio of 0.8:0.2 among 2000 sample instances) considering the default hyperparameter settings for other methods. For NLPOpt-Net, the parameters are $\alpha=10$ and $k=1$ if not mentioned otherwise. All the problem generation descriptions are given in Appendix \ref{problem_generation}.

\subsection{Quadratic Programs (QPs)}\label{results_qp}
We consider solving convex quadratic programs with a quadratic objective and affine constraints:
\begin{ceqn}
\begin{equation}\label{qp_form}
\begin{aligned}
\min_{y} \quad & \frac{1}{2} y^\top Q y + c^\top y \ ,\text{s.t.}\ A y = b + B x,\ C y \leq d,\ l + L x \leq y \leq u + U x,
\end{aligned}
\end{equation}
\end{ceqn}
where, $Q\in\mathbb{R}^{n\times n}\succeq0$, $c\in\mathbb{R}^n$, $A\in\mathbb{R}^{n_\textrm{eq}\times n}$, $b\in\mathbb{R}^{n_\textrm{eq}}$, $B\in\mathbb{R}^{n_\textrm{eq}\times p}$, $C\in\mathbb{R}^{n_\textrm{ineq}\times n}$, $d\in\mathbb{R}^{n_\textrm{ineq}}$, $l\in\mathbb{R}^n$, $L\in\mathbb{R}^{n\times p}$, $u\in\mathbb{R}^n$, and $U\in\mathbb{R}^{n\times p}$. The goal is to find optimal $y\in\mathbb{R}^n$ for different parameter realizations $x\in\mathbb{R}^p$. Table \ref{tab:Results} compares the performance of NLPOpt-Net with other methods for different cases. We find that both NLPOpt-Net and DC3 preserves feasibility. However, the maximum inequality violation is not zero for DC3 indicating that for a very small portion of instances DC3 is unable to preserve feasibility whereas NLPOpt-Net shows no infeasibility across different runs. Soft constrained NN includes infeasibility and goes below the true objectives. Eq-NN model preserves feasibility in terms of equality constraints, but not for inequality constraints. In terms of computational time, NLPOpt-Net requires same order of magnitude time for this case as DC3. In addition to the feasibility assurance, a major improvement we find is in terms of optimality. While DC3 has the average optimality gap of more than $150$\% after training for 2000 epochs, NLPOpt-Net finds exact optimal solutions when rounded up to three decimals. 
\begin{table}[h]
\caption{Results for parameterized optimization problems in terms of objective, constraint violation and time. The standard deviations across different runs are reported in the parenthesis. After training, both training and validation instances are tested. Values are rounded off to three decimals.\\}
\label{tab:Results}
\centering
\resizebox{\textwidth}{!}{%
\begin{tabular}{lrrrrrr}
\toprule
 & Avg. Obj.  & Max eq. & Mean eq. & Max ineq. & Mean ineq. & Time (s/batch) \\
\midrule
\multicolumn{7}{c}{Convex QP, $\alpha$=10, $k$=1}\\\midrule
OSQP & -8.730 (0.000) & 0.000 (0.000) & 0.000 (0.000) & 0.000 (0.000) & 0.000 (0.000) & 0.0039 (0.0000) \\
NN & -18.100 (0.055) & 0.631 (0.000) & 0.161 (0.000) & 0.005 (0.000) & 0.000 (0.000) & 0.0052 (0.0000) \\
Eq. NN & 5.132 (0.187) & 0.000 (0.000) & 0.000 (0.000) & 2.356 (0.084) & 0.003 (0.000) & 0.0058 (0.0000) \\
DC3 & -3.213 (1.441) & 0.000 (0.000) & 0.000 (0.000) & 0.607 (0.190) & 0.000 (0.000) & 0.012 (0.0000) \\
NLPOpt-Net & -8.730 (0.000) & 0.000 (0.000) & 0.000 (0.000) & 0.000 (0.000) & 0.000 (0.000) & 0.0357 (0.0000) \\
\midrule
\multicolumn{7}{c}{Convex QCQP, $\alpha=10$, $k=5$}\\\midrule
SCS & 3.227 (0.055) & 0.000 (0.000) & 0.000 (0.000) & 0.000 (0.000) & 0.000 (0.000) & 0.0514 (0.0000) \\
NN & 2.583 (0.161) & 0.210 (0.110) & 0.033 (0.000) & 0.000 (0.000) & 0.000 (0.000) & 0.0070 (0.025) \\
Eq. NN & 17.270 (0.929) & 0.000 (0.000) & 0.000 (0.000) & 2.276 (0.699) & 0.012 (0.045) & 0.0080 (0.032) \\
DC3 & 7.186 (0.646) & 0.000 (0.000) & 0.000 (0.000) & 0.306 (0.167) & 0.000 (0.000) & 0.0240 (0.035) \\
NLPOpt-Net & 3.227 (0.000) & 0.000 (0.000) & 0.000 (0.000) & 0.000 (0.000) & 0.000 (0.000) & 0.4260 (0.020) \\
\midrule
\multicolumn{7}{c}{Convex NLP, $\alpha=100$, $k=5$}\\\midrule
SCS & 113.300 (0.000) & 0.000 (0.000) & 0.000 (0.000) & 0.003 (0.000) & 0.000 (0.000) & 0.2510 (0.0000) \\
NN & 113.200 (0.336) & 0.017 (0.032) & 0.003 (0.000) & 0.378 (0.474) & 0.000 (0.000) & 0.0070 (0.010) \\
Eq. NN & 296.600 (5.916) & 0.000 (0.000) & 0.000 (0.000) & 18.250 (1.593) & 2.517 (0.684) & 0.0070 (0.010) \\
DC3 & 114.600 (0.677) & 0.000 (0.000) & 0.000 (0.000) & 1.540 (0.639) & 0.000 (0.000) & 0.0280 (0.014) \\
NLPOpt-Net & 113.300 (0.000) & 0.000 (0.000) & 0.000 (0.000) & 0.000 (0.000) & 0.000 (0.000) & 0.4300 (0.036) \\
\midrule
\multicolumn{7}{c}{Nonconvex NLP, $\alpha=10$, $k=1$}\\\midrule
SLSQP & -3.833 (0.000) & 0.000 (0.000) & 0.000 (0.000) & 0.000 (0.000) & 0.000 (0.000) & 0.0638 (0.0000) \\
NN & -4.098 (0.110) & 0.853 (0.257) & 0.177 (0.000) & 0.000 (0.000) & 0.000 (0.000) & 0.0040 (0.010) \\
Eq. NN & -1.306 (1.008) & 0.000 (0.000) & 0.000 (0.000) & 5.436 (1.347) & 0.022 (0.122) & 0.0050 (0.010) \\
DC3 & -2.803 (0.736) & 0.000 (0.000) & 0.000 (0.000) & 0.961 (1.043) & 0.000 (0.000) & 0.0110 (0.020) \\
NLPOpt-Net & -3.832 (0.000) & 0.000 (0.000) & 0.000 (0.000) & 0.000 (0.000) & 0.000 (0.000) & 0.0780 (0.022) \\
\bottomrule
\end{tabular}
}
\end{table}
\subsection{Quadratically Constrained Quadratic Programs (QCQPs)}\label{results_qcqp}
QCQPs are required to solve in many real life scenario, such as portfolio optimization \cite{boyd2004convex}. In this example, we test the performance of NLPOpt-Net for the following form of parameterized convex QCQP:
\begin{ceqn}
\begin{equation}\label{qcqp_form}
\begin{aligned}
\min_{y}\ \frac{1}{2} y^\top Q y + c^\top \ , \text{s.t.} \ A y = b + B x, \ y^\top C_i y + d_i^\top y \leq \beta_i + E_i x, \ l + L x \leq y \leq u + U x.
\end{aligned}
\end{equation}
\end{ceqn}
where, $i = 1, 2, \ldots, n_\mathrm{ineq}$. For constants, $Q \in \mathbb{R}^{n \times n} \succeq 0$, $c \in \mathbb{R}^n$, $A \in \mathbb{R}^{n_{\mathrm{eq}} \times n}$, $b \in \mathbb{R}^{n_{\mathrm{eq}}}$, $B \in \mathbb{R}^{n_{\mathrm{eq}} \times p }$, $C_i \in \mathbb{R}^{n \times n} \succeq 0$, $d_i \in \mathbb{R}^n$, $\beta_i \in \mathbb{R}$, $E_i \in \mathbb{R}^{1 \times p}$ for $i = 1, 2, \ldots, n_{\mathrm{ineq}}$, $l \in \mathbb{R}^n$, $L \in \mathbb{R}^{n \times p}$, $u \in \mathbb{R}^n$, and $U \in \mathbb{R}^{n \times p}$, we must learn to approximate the optimal solution $y \in \mathbb{R}^n$ given a parameter realization $x \in \mathbb{R}^p$. As of Table \ref{tab:Results} we find the behavior of feasibility for QCQP similar to QP case. While the mean inequality violation is zero for DC3, it struggles for a very few instance. However, for this test case, NLPOpt-Net requires one order of magnitude more time than DC3 and the optimizer. Although the work is done in batch and the optimizer time is taken as a average time for solving instances over a batch to ensure complete parallelization, the time required is around 9 times slower than the optimizer due to the number of layers ($k$=5) used in projection layer. Moreover, the best objectives are found for NLPOpt-Net across all the learning methods providing near optimal solution with feasibility.
\subsection{Convex Nonlinear Programs (NLPs)}\label{results_nlp}
In this case we consider a general nonlinear convex constraint and kept the other structure same as QP. we consider exponential terms in inequality constraints making them nonlinear. We use the following form of convex nonlinear problem.
\begin{ceqn}
\small
\begin{equation}\label{nlp_form}
\begin{aligned}
\min_{y} \ \frac{1}{2} y^\top Q y + c^\top y\ , \text{s.t.}\ A y = b + B x, \ a_i^\top \exp(y) + y^\top W y \leq \beta_i + E_i x, \ l + L x \leq y \leq u + U x.
\end{aligned}
\end{equation}
\end{ceqn}
where, $i = 1, 2, \ldots, n_\mathrm{ineq}$. For constants, $Q \in \mathbb{R}^{n \times n} \succeq 0$, $c \in \mathbb{R}^n$, $A \in \mathbb{R}^{n_{\textrm{eq}} \times n}$, $b \in \mathbb{R}^{n_{\textrm{eq}}}$, $B \in \mathbb{R}^{n_{\textrm{eq}} \times p}$, $a_i \in \mathbb{R}^n$, $W \in \mathbb{R}^{n \times n} \succeq 0$, $\beta_i \in \mathbb{R}$, $E_i \in \mathbb{R}^{1 \times p}$ for $i = 1,2,\ldots,n_{\textrm{ineq}}$, $l \in \mathbb{R}^n$, $L \in \mathbb{R}^{n \times p}$, $u \in \mathbb{R}^n$, and $U \in \mathbb{R}^{n \times p}$. As of Table \ref{tab:Results}, for $\alpha=100$ and $k=5$, NLPOpt-Net achieves near optimal solution in compared to all the other learning methods and also retains complete feasibility across all the instances. For this case DC3 also achieves a near optimal solution.
\subsection{Simple Nonconvex Programs}\label{results_nonconvex}
We now consider the simple non-convex introduced by Donti et al.\cite{donti2021dc3}. The problem is defined as following:
\begin{ceqn}
\begin{equation}\label{nonconvex_form}
\begin{aligned}
\min_{y} \quad & \frac{1}{2} y^\top Q y + p^\top \sin(y) \quad \text{s.t.} \quad A y = x,\ G y \leq h,
\end{aligned}
\end{equation}
\end{ceqn}
where, $Q\in\mathbb{R}^{n\times n}\succeq0$, $p\in\mathbb{R}^n$, $A\in\mathbb{R}^{n_\textrm{eq}\times n}$, $G\in\mathbb{R}^{n_\textrm{ineq}\times n}$ and $h\in\mathbb{R}^{n_\textrm{ineq}}$. $\sin(y)$ denotes componentwise application of the sine function to the vector $y$. We generated the problems using the similar steps followed in \cite{donti2021dc3}. We consider \texttt{SLSQP} as the optimizer for this nonconvex case. Even though the NLPOpt-Net framework is designed for convex cases we test the nonconvex case where NLPOpt-Net achieved best optimality among all the learning methods (see Table \ref{tab:Results}, optimality gap $\approx 0.026\%$). Due to the constraints being affine, a single layer in the projection is sufficient to satisfy the constraints resulting into efficient inference time. Moreover, the optimality doesn't varies across the runs where as DC3 has  $\approx 19\%$ variation in different runs.

\section{Conclusions}\label{conclusions}
We present NLPOpt-Net, a neural network architecture that learns parameter to solution mapping of parametric optimization problem with feasibility guarantee in the constraint manifold. Our approach includes a $k$-layered projection step where $k$ can vary to achieve feasibility. We show that the projection layer has a descent property under simple assumptions therefore achieving no worse projection in terms of optimality. In our $k$-layered projection, we employ a case specific version of Chambolle-Pock algorithm to solve a structured quadratic program with efficiency $k$ times to retain feasibility. While the forward pass is an inversion free method, we leverage on the implicit function theorem to provide implicit backward gradient with a custom vjp (Vector Jacobian Product) route in JAX. We find that NLPOpt-Net yield superor results in terms of optimality with better feasibility than other deep learning method based solvers. In general, NLPOpt-Net is designed for convex problems, we find that in some specific cases, it may yield superior result for nonconvex problems too. Furthermore, we show NLPOpt-Net preserves the correct active set in the constraints leading towards better optimality. Depending on the number of the layers in projection, the computational expense and feasibility may vary. While the framework is broadly applicable, it will be possible to design further improvement with specific constraint structure and problem instances. Additionally, sparsity pattern in the constraints like real life examples, this framework can be utilized with efficient matrix vector multiplication to handle large problems.
\section*{Acknowledgment} 
The authors gratefully acknowledge partial financial support from the U.S. National Science Foundation (NSF CAREER award CBET-1943479) and the US EPA Project Grant
84097201.

\bibliographystyle{ieeetr} 
\begingroup
\setstretch{1.0}
\bibliography{References}
\endgroup

\newpage

\appendix
\section{Problem instance generation}\label{problem_generation}
We provide the details of the problem generation for different computational studies here:
\paragraph{QP:}We consider $n=100,p=50, n_\textrm{eq}=50, n_\textrm{ineq}=50$, and the parameter bounds are taken as $x\in[-10,10]^{50}$. During training, the parameter entries were drawn from the uniform distribution on $[-10,10]$ and we consider 2000 problem instances. For the $Q$ matrix, we generate a random Gaussian matrix $M$ and consider $Q=(M^\top M)/n+ \textrm{diag}(\delta)$. We add the diagonal shift $\delta$ between $[1.2,1.8]$ ensuring $Q$ is positive definite, so the objective is always strictly convex. The vector $c$ is created elementwise and all the entries are drawn from the uniform distribution on $[0.5,1.5]$. The matrix $A$ is generated from a QR factorization of a random Gaussian matrix, where the first $n_\textrm{eq}$ orthonormal directions are selected and scaled row-wise to ensure full row rank and good numerical conditioning. To ensure feasibility of the equality constraints for all parameter realizations, $x\in[x_L,x_U]$, we construct an affine mapping of the form $y_f(x)=y_c+Tx$, where $y_c\in\mathbb{R}^n$ and $T\in\mathbb{R}^{n\times p}$ that are randomly generated with controlled magnitude. Then other constant vector and matrix for equality constraints are defined as $b=Ay_c$ and $B=AT$. For inequality constraints, each row of matrix $C$ is generated with sparse support and normalized coefficients. Vector $d$ is defined as $d_i=C_i y_c+\max_{x\in[x_L,x_U]}C_iTx$ which guarantees feasibility for all admissible $x$. Finally, the bound constraints are defined with $L=U=T$, $l=y_c-rI$, and $u=y_c+rI$ for some $r>0$. This ensures feasibility of the constraints for any realization of $x$ within the bounds.
\paragraph{QCQP:}The objective and equality constraints are generated in same way as the quadratic case. For quadratic inequality constraints, we construct each constraint in a structured manner. Specifically, for each $i$, we generate a direction vector $r_i\in\mathbb{R}^n$ and define $C_i=\alpha_i r_i r_i^\top$ and $d_i=r_i$, where $\alpha_i>0$ is chosen to control the curvature and ensure numerical stability. Thus, each constraint takes the form
\begin{ceqn}
\begin{equation}
\frac{1}{2}\alpha_i (r_i^\top y)^2 + r_i^\top y \leq \beta_i + E_i x.
\end{equation}
\end{ceqn}
The right-hand side is constructed to guarantee feasibility of the affine mapping $y_f(x)$ for all admissible $x$. In particular, $\beta_i$ is chosen as
\begin{ceqn}
\begin{equation}
\beta_i = \max_{x\in[x_L,x_U]} 
\left( \frac{1}{2}\alpha_i (r_i^\top y_f(x))^2 + r_i^\top y_f(x) - E_i x \right),
\end{equation}
\end{ceqn}
which ensures that $y_f(x)$ satisfies all quadratic inequalities over the parameter domain. The parameter-dependent term $E_i x$ is introduced by selecting $E_i$ along representative directions in the parameter space, enabling nontrivial coupling between $x$ and the constraints. Finally, the bound constraints are defined with $L=U=T$ and $l=y_c-rI$ and $u=y_c+rI$ for some $r>0$. This ensures feasibility of the constraints for any realization of $x$ within the bound.
\paragraph{Convex NLP:} For nonlinear inequality constraints, we construct each constraint in a structured manner. Specifically, for each $i$, we generate a vector $a_i \in \mathbb{R}^n$ and a positive semi-definite matrix $W_i \in \mathbb{R}^{n \times n}$ with sparse diagonal structure. The nonlinear constraint takes the form
\begin{ceqn}
\begin{equation}
a_i^\top \exp(y) + y^\top W_i y \leq \beta_i + E_i x.
\end{equation}
\end{ceqn}
The coefficients $a_i$ and $W_i$ are constructed by selecting a small subset of variables and assigning nonzero weights to emphasize nonlinear behavior along those directions. The right-hand side $\beta_i$ is defined as
\begin{ceqn}
\begin{equation}
\beta_i = \max_{x\in[x_L,x_U]} 
\left(a_i^\top \exp(y_f(x)) + y_f(x)^\top W_i y_f(x) - E_i x\right) + \epsilon,
\end{equation}
\end{ceqn}
where $\epsilon > 0$ is a margin parameter. This ensures that the affine mapping $y_f(x)$ satisfies all nonlinear inequality constraints for every $x$ in the parameter domain. The parameter-dependent term $E_i x$ is constructed by selecting directions in the parameter space and scaling them to introduce coupling between $x$ and the nonlinear constraints. Rest of the vectors and matrices are generated as per previous instructions.
\paragraph{Nonconvex:}We used the methods provided by Donti et. al.\cite{donti2021dc3} to generate the problems.

\section{Additional experiments}\label{additional}


In the additional experiments we compare NLPOpt-Net with Optimizer and DC3 for different settings on the QPs. In the additional experiments, we run QP with 100 variables as the number of equality constraint (Table \ref{tab:equality_comparison_stacked}), inequality constraints (Table \ref{tab:inequality_comparison_stacked}), number of parameters (Table \ref{tab:parameter_comparison_stacked}) and training sample size varies (Table \ref{tab:training_fraction_comparison_stacked}). Additionally we provide the active set agreement analysis for different settings. This shows how from the dual variables we can detect the active sets. For inequality constraint set $I_\textrm{ineq}$, we consider a constraint $g_i \,\forall i\in I_\textrm{ineq}$ is active when the dual variable $\mu_i>0$ and we denote the set of active constraints as $I_\textrm{A,ineq}$. We compare across all the runs and instances by one to one mapping, what fraction of the active sets were predicted correctly with the true active sets from the optimizer run. The activity agreements are summarized in Table \ref{tab:active_set_stacked_performance}. We take the default model specific hyperparameters for DC3 and the hyperparameter tuning may yield better results in terms of optimality and constraint violation. We train all the models for 2000 epochs. We observe that DC3 objective was still improving. Providing more training epoch may yield better objective with less variation. For NLPOpt-Net the only hyperparameter we consider for all the cases is weight for consistency loss ($\alpha$).

\textit{\textbf{Remarks on Additional Experiments}.} (see Appendix \ref{additional}) We find that the time requirement for NLPOpt-Net during inference increases with equality constraints and sometimes with inequality constraints. The underlying reason could be the sparsity pattern of the constraint matrices. As the Chambolle-Pock algorithm rely on matrix multiplications in each step and on norm calculation, the denser the matrix the more time it will take. In our problems we generated fully dense matrix. In practical setting constraints are sparse and techniques can be employed to make the matrix multiplications efficient by detecting the sparsity pattern. Grontas et al. \cite{grontas2025pinet} shows such case where by considering sparsity similar first order methods can be used to deal with very large problems. We find from Table \ref{tab:training_fraction_comparison_stacked}, NLPOpt-Net deviates from optimal solution very steadily when the training sample is smaller while DC3 struggles to find a good solution when the training samples are limited. To the best of our knowledge, only NLPOpt-Net provides idea on the active sets of the constraints and Table \ref{tab:active_set_stacked_performance} shows a near accurate prediction of active sets.
\begin{table}[h]
\caption{Results on QP task for 100 variables, 50 equality and 50 inequality constraints as the number of parameters varies as 10, 25, 50, 75, 100. We compare the performance of NLPOpt-Net with DC3 and the optimizer according to objective value, maximum/mean equality and inequality violation and inference time per batch. Standard deviation across 5 runs are shown in parenthesis.\\}
\label{tab:parameter_comparison_stacked}
\centering
\resizebox{\textwidth}{!}{%
\begin{tabular}{llccccc}
\toprule
Model & Metric & 10 & 25 & 50 & 75 & 100 \\
\midrule
Optimizer & Obj. val. & -8.026 (0.089) & -8.352 (0.089) & -8.730 (0.000) & -8.471 (0.000) & -8.892 (0.000) \\
 & Max eq. & 0.000 (0.000) & 0.000 (0.000) & 0.000 (0.000) & 0.000 (0.000) & 0.000 (0.000) \\
 & Mean eq. & 0.000 (0.000) & 0.000 (0.000) & 0.000 (0.000) & 0.000 (0.000) & 0.000 (0.000) \\
 & Max ineq. & 0.000 (0.000) & 0.000 (0.000) & 0.000 (0.000) & 0.000 (0.000) & 0.000 (0.000) \\
 & Mean ineq. & 0.000 (0.000) & 0.000 (0.000) & 0.000 (0.000) & 0.000 (0.000) & 0.000 (0.000) \\
 & Time (s/sample) & 0.004 (0.000) & 0.004 (0.000) & 0.004 (0.000) & 0.004 (0.000) & 0.003 (0.000) \\
 \midrule
DC3 & Obj. val. & -1.536 (1.770) & 1.082 (3.576) & -3.213 (1.441) & -3.364 (1.025) & -2.652 (1.777) \\
 & Max eq. & 0.000 (0.000) & 0.000 (0.000) & 0.000 (0.000) & 0.000 (0.000) & 0.000 (0.000) \\
 & Mean eq. & 0.000 (0.000) & 0.000 (0.000) & 0.000 (0.000) & 0.000 (0.000) & 0.000 (0.000) \\
 & Max ineq. & 0.988 (0.228) & 1.673 (0.837) & 0.607 (0.190) & 0.696 (0.302) & 1.468 (0.219) \\
 & Mean ineq. & 0.001 (0.000) & 0.001 (0.001) & 0.000 (0.000) & 0.000 (0.000) & 0.001 (0.000) \\
 & Time (s/batch) & 0.014 (0.000) & 0.013 (0.000) & 0.012 (0.000) & 0.013 (0.000) & 0.014 (0.000) \\
\midrule
NLPOpt-Net & Obj. val. & -8.024 (0.000) & -8.350 (0.000) & -8.730 (0.000) & -8.468 (0.000) & -8.888 (0.000) \\
 & Max eq. & 0.000 (0.000) & 0.000 (0.000) & 0.000 (0.000) & 0.000 (0.000) & 0.000 (0.000) \\
 & Mean eq. & 0.000 (0.000) & 0.000 (0.000) & 0.000 (0.000) & 0.000 (0.000) & 0.000 (0.000) \\
 & Max ineq. & 0.000 (0.000) & 0.000 (0.000) & 0.000 (0.000) & 0.000 (0.000) & 0.000 (0.000) \\
 & Mean ineq. & 0.000 (0.000) & 0.000 (0.000) & 0.000 (0.000) & 0.000 (0.000) & 0.000 (0.000) \\
 & Time (s/batch) & 0.036 (0.000) & 0.036 (0.000) & 0.036 (0.000) & 0.035 (0.000) & 0.033 (0.000) \\
\bottomrule
\end{tabular}
}
\end{table}

\begin{table}[h]
\caption{Results on QP task for 100 variables, 50 parameters and 50 inequality constraints as the number of equality constraints varies as 10, 30, 50, 70, 90.}
\label{tab:equality_comparison_stacked}
\centering
\resizebox{\textwidth}{!}{%
\begin{tabular}{llccccc}
\toprule
Model & Metric & 10 & 30 & 50 & 70 & 90 \\
\midrule
Optimizer & Obj. val. & -8.772 (0.000) & -8.730 (0.000) & -8.730 (0.000) & -8.365 (0.000) & 3.814 (0.000) \\
 & Max eq. & 0.000 (0.000) & 0.000 (0.000) & 0.000 (0.000) & 0.000 (0.000) & 0.000 (0.000) \\
 & Mean eq. & 0.000 (0.000) & 0.000 (0.000) & 0.000 (0.000) & 0.000 (0.000) & 0.000 (0.000) \\
 & Max ineq. & 0.000 (0.000) & 0.000 (0.000) & 0.000 (0.000) & 0.000 (0.000) & 0.000 (0.000) \\
 & Mean ineq. & 0.000 (0.000) & 0.000 (0.000) & 0.000 (0.000) & 0.000 (0.000) & 0.000 (0.000) \\
 & Time (s/sample) & 0.003 (0.000) & 0.004 (0.000) & 0.004 (0.000) & 0.004 (0.000) & 0.004 (0.000) \\
\midrule
DC3 & Obj. val. & -3.757 (1.141) & -3.288 (1.361) & -3.213 (1.441) & -2.177 (1.894) & 4.102 (1.928) \\
 & Max eq. & 0.000 (0.000) & 0.000 (0.000) & 0.000 (0.000) & 0.000 (0.000) & 0.000 (0.000) \\
 & Mean eq. & 0.000 (0.000) & 0.000 (0.000) & 0.000 (0.000) & 0.000 (0.000) & 0.000 (0.000) \\
 & Max ineq. & 0.395 (0.141) & 0.557 (0.195) & 0.607 (0.190) & 0.977 (0.319) & 1.123 (0.594) \\
 & Mean ineq. & 0.000 (0.000) & 0.000 (0.000) & 0.000 (0.000) & 0.001 (0.000) & 0.001 (0.000) \\
 & Time (s/batch) & 0.013 (0.000) & 0.013 (0.000) & 0.012 (0.000) & 0.013 (0.000) & 0.013 (0.000) \\
\midrule
NLPOpt-Net & Obj. val. & -8.772 (0.000) & -8.730 (0.000) & -8.730 (0.000) & -8.365 (0.000) & 3.819 (0.000) \\
 & Max eq. & 0.000 (0.000) & 0.000 (0.000) & 0.000 (0.000) & 0.000 (0.000) & 0.000 (0.000) \\
 & Mean eq. & 0.000 (0.000) & 0.000 (0.000) & 0.000 (0.000) & 0.000 (0.000) & 0.000 (0.000) \\
 & Max ineq. & 0.000 (0.000) & 0.000 (0.000) & 0.000 (0.000) & 0.000 (0.000) & 0.000 (0.000) \\
 & Mean ineq. & 0.000 (0.000) & 0.000 (0.000) & 0.000 (0.000) & 0.000 (0.000) & 0.000 (0.000) \\
 & Time (s/batch) & 0.018 (0.000) & 0.027 (0.000) & 0.036 (0.000) & 0.040 (0.000) & 0.045 (0.000) \\
\bottomrule
\end{tabular}
}
\end{table}

\begin{table}[h]
\caption{Results on QP task for 100 variables, 50 parameters and 50 equality constraints as the number of inequality constraints varies as 10, 30, 50, 70, 90.}
\label{tab:inequality_comparison_stacked}
\centering
\resizebox{\textwidth}{!}{%
\begin{tabular}{llccccc}
\toprule
Model & Metric & 10 & 30 & 50 & 70 & 90 \\
\midrule
Optimizer & Obj. val. & -19.310 (0.000) & -11.020 (0.000) & -8.730 (0.000) & -6.551 (0.084) & -8.363 (0.063) \\
 & Max eq. & 0.000 (0.000) & 0.000 (0.000) & 0.000 (0.000) & 0.000 (0.000) & 0.000 (0.000) \\
 & Mean eq. & 0.000 (0.000) & 0.000 (0.000) & 0.000 (0.000) & 0.000 (0.000) & 0.000 (0.000) \\
 & Max ineq. & 0.000 (0.000) & 0.000 (0.000) & 0.000 (0.000) & 0.000 (0.000) & 0.000 (0.000) \\
 & Mean ineq. & 0.000 (0.000) & 0.000 (0.000) & 0.000 (0.000) & 0.000 (0.000) & 0.000 (0.000) \\
 & Time (s/sample) & 0.002 (0.000) & 0.003 (0.000) & 0.004 (0.000) & 0.005 (0.000) & 0.005 (0.000) \\
 \midrule
DC3 & Obj. val. & -3.757 (1.141) & -3.288 (1.361) & -3.213 (1.441) & -2.177 (1.894) & -2.119 (1.928) \\
 & Max eq. & 0.000 (0.000) & 0.000 (0.000) & 0.000 (0.000) & 0.000 (0.000) & 0.000 (0.000) \\
 & Mean eq. & 0.000 (0.000) & 0.000 (0.000) & 0.000 (0.000) & 0.000 (0.000) & 0.000 (0.000) \\
 & Max ineq. & 0.395 (0.141) & 0.557 (0.195) & 0.607 (0.190) & 0.977 (0.319) & 1.123 (0.594) \\
 & Mean ineq. & 0.000 (0.000) & 0.000 (0.000) & 0.000 (0.000) & 0.001 (0.000) & 0.001 (0.000) \\
 & Time (s/batch) & 0.013 (0.000) & 0.013 (0.000) & 0.012 (0.000) & 0.013 (0.000) & 0.013 (0.000) \\
\midrule
NLPOpt-Net & Obj. val. & -19.310 (0.000) & -11.020 (0.000) & -8.730 (0.000) & -6.548 (0.000) & -8.363 (0.000) \\
 & Max eq. & 0.000 (0.000) & 0.000 (0.000) & 0.000 (0.000) & 0.000 (0.000) & 0.000 (0.000) \\
 & Mean eq. & 0.000 (0.000) & 0.000 (0.000) & 0.000 (0.000) & 0.000 (0.000) & 0.000 (0.000) \\
 & Max ineq. & 0.000 (0.000) & 0.000 (0.000) & 0.000 (0.000) & 0.000 (0.000) & 0.000 (0.000) \\
 & Mean ineq. & 0.000 (0.000) & 0.000 (0.000) & 0.000 (0.000) & 0.000 (0.000) & 0.000 (0.000) \\
 & Time (s/batch) & 0.019 (0.000) & 0.026 (0.000) & 0.036 (0.000) & 0.044 (0.000) & 0.026 (0.000) \\
\bottomrule
\end{tabular}
}
\end{table}

\begin{table}[h]
\caption{Results on QP task for 100 variables, 50 parameters, 50 inequality constraints and 50 parameters as the number of samples taken for training is varying. We consider total 2000 instances and for training we vary the training fraction as 0.8, 0.6, 0.4, 0.2.}
\label{tab:training_fraction_comparison_stacked}
\centering
\resizebox{\textwidth}{!}{%
\begin{tabular}{llcccc}
\toprule
Model & Metric & 0.8 & 0.6 & 0.4 & 0.2 \\
\midrule
Optimizer & Objective & -8.730 (0.000) & -8.730 (0.095) & -8.730 (0.095) & -8.730 (0.095) \\
 & Max. Eq. & 0.000 (0.000) & 0.000 (0.000) & 0.000 (0.000) & 0.000 (0.000) \\
 & Mean Eq. & 0.000 (0.000) & 0.000 (0.000) & 0.000 (0.000) & 0.000 (0.000) \\
 & Max. Ineq. & 0.000 (0.000) & 0.000 (0.000) & 0.000 (0.000) & 0.000 (0.000) \\
 & Mean Ineq. & 0.000 (0.000) & 0.000 (0.000) & 0.000 (0.000) & 0.000 (0.000) \\
 & Time (s/sample) & 0.004 (0.000) & 0.004 (0.000) & 0.004 (0.000) & 0.004 (0.000) \\
 \midrule
DC3 & Objective & -3.213 (0.845) & -0.922 (1.348) & 3.189 (1.478) & 23.650 (2.317) \\
 & Max. Eq. & 0.000 (0.000) & 0.000 (0.000) & 0.000 (0.000) & 0.000 (0.000) \\
 & Mean Eq. & 0.000 (0.000) & 0.000 (0.000) & 0.000 (0.000) & 0.000 (0.000) \\
 & Max. Ineq. & 0.607 (0.276) & 1.132 (0.656) & 1.099 (0.409) & 3.201 (0.770) \\
 & Mean Ineq. & 0.000 (0.000) & 0.000 (0.000) & 0.001 (0.000) & 0.006 (0.045) \\
 & Time (s/batch) & 0.012 (0.000) & 0.014 (0.032) & 0.015 (0.045) & 0.015 (0.045) \\
\midrule
NLPOpt-Net & Objective & -8.730 (0.000) & -8.726 (0.000) & -8.662 (0.095) & -8.271 (0.130) \\
 & Max. Eq. & 0.000 (0.000) & 0.000 (0.000) & 0.000 (0.000) & 0.000 (0.000) \\
 & Mean Eq. & 0.000 (0.000) & 0.000 (0.000) & 0.000 (0.000) & 0.000 (0.000) \\
 & Max. Ineq. & 0.000 (0.000) & 0.000 (0.000) & 0.000 (0.000) & 0.000 (0.000) \\
 & Mean Ineq. & 0.000 (0.000) & 0.000 (0.000) & 0.000 (0.000) & 0.000 (0.000) \\
 & Time (s/batch) & 0.035 (0.000) & 0.039 (0.000) & 0.034 (0.000) & 0.027 (0.000) \\
\bottomrule
\end{tabular}
}
\end{table}
\begin{table}[h]
\centering
\caption{Performance across problem classes and parameter settings. The first row provides the active set agreement for the results shown in Section \ref{ComStd}. Other rows shows results on QP with 100 variables and the common structure with 50 equality, 50 inequality and 50 parameters with varying one of the property at a time. We cap the dual variables to $1\times 10^{-6}$}
\label{tab:active_set_stacked_performance}
\resizebox{\textwidth}{!}{
\begin{tabular}{llccccc}
\toprule
\multirow{2}{*}{Class} & Problem type & QP (\ref{results_qp}) & QCQP (\ref{results_qcqp}) & NLP (\ref{results_nlp}) & Nonconvex (\ref{results_nonconvex}) & {} \\
& Active set agreement & 0.998 & 0.997 & 0.999 & 0.999 & {} \\\midrule
\multirow{2}{*}{$p$ (QP)} & No. of parameters & 10 & 25 & 50 & 75 & 10 \\
& Active set agreement & 0.998 & 0.998 & 0.998 & 0.999 & 0.998 \\ \midrule
\multirow{2}{*}{$n_\textrm{eq}$ (QP)} & No. of equality & 10 & 30 & 50 & 70 & 90 \\
& Active set agreement & 0.997 & 0.997 & 0.998 & 0.999 & 1.000 \\ \midrule
\multirow{2}{*}{$n_\textrm{ineq}$ (QP)} & No. of inequality & 10 & 30 & 50 & 70 & 90 \\
& Active set agreement & 0.997 & 0.998 & 0.998 & 0.998 & 0.998 \\
\bottomrule
\end{tabular}}
\end{table}

\section{Local convergence rate of the projection layer}\label{local_convergence}
The $k$-layer projection is an iterative procedure to polish the solution to get back feasibility. This can be seen as a similar technique like sequential quadratic programming (SQP). However, we do not imply the second order information of the Lagrangian. Instead we refine the quadratic objective. The convergence rate of the method can be analyzed from SQP theory\cite{nocedal2006numerical}. A similar type of convergence analysis is discussed for nonlinear equality constraints in AdaNP Projection\cite{lastrucci2025enforce}.\\

The $k$-layer projection at the $k$th layer ($\mathcal{P}_s^k$) solves the subproblem of the form:

\begin{ceqn}
\begin{equation}\label{p_approx_conv}\tag{$\mathrm{P}_\mathrm{approx}$}
    \begin{aligned}
        \min_{{y}\in\mathcal{B}(x)} \quad & f({x},\hat{{y}}) + J^\top({y}-\hat{{y}}) + \frac{1}{2}({y}-\hat{{y}})^\top H_{d}({y}-\hat{{y}})\\
        \mathrm{s.t.} \quad   & J_h(x,y^k)(y-y^k)+h(x,y^k)=0\\
                            & J_g(x,y^k)(y-y^k)+g(x,y^k)\le0
    \end{aligned}
\end{equation}
\end{ceqn}

which approximates the original nonlinear problem:

\begin{ceqn}
\begin{equation}\label{p_orig_conv}\tag{$\mathrm{P}_\mathrm{orig}$}
    \begin{aligned}
        \min_{{y}\in\mathcal{B}(x)} \quad & f({x},\hat{{y}}) + J^\top({y}-\hat{{y}}) + \frac{1}{2}({y}-\hat{{y}})^\top H_{d}({y}-\hat{{y}})\\
        \mathrm{s.t.} \quad   & h(x,y)=0\\
                            & g(x,y)\le0
    \end{aligned}
\end{equation}
\end{ceqn}

where, $\mathcal{B}$ defines the box constraint on the decision variables. While the hessian of the approximated objective includes the hessian of the Lagrangian, we keep original objective as it is quadratic already. As we discussed, the original problem is convex, therefore we assume the existence of $y^*$, a global solution to the original problem at which the following conditions (\ref{assumpstart_conv}-\ref{assumpend_conv}) hold:

\begin{enumerate}[label=C\arabic*]\label{assump_C}
    \item The objective function and the constraints are twice differentiable in a neighborhood of $y^*$ with Lipschitz continuous second order derivatives. \label{assumpstart_conv}
    \item Slater's condition holds at $y^*$, therefore, the KKT conditions are satisfied at $y^*$ with a valid set of Lagrangian multipliers $\lambda^*$ (equality) and $\mu^*$ (inequality) with strong duality.
    \item Second-order sufficient conditions hold at ($y^*,\lambda^*,\mu^*$). \label{assumpend_conv}
\end{enumerate}

Let $\mathcal{L}_\mathrm{orig}$ is the Lagrangian of the original problem (\ref{p_orig_conv}). With slack variables and Fischer-Burmeister reformulation the KKT system can be expressed as (see \cite{iftakher2025physics} for details on the formulation):

\begin{ceqn}
    \begin{equation}
        F(z) = \begin{bmatrix}
            \nabla_y\mathcal{L}_\mathrm{orig}(y,\lambda,\mu)\\
            h(y)\\
            g(y)+s\\
            \phi(\mu,s)
        \end{bmatrix}, \quad \mathrm{with} \quad z = \begin{bmatrix}
            y\\\lambda\\\mu\\s
        \end{bmatrix},
    \end{equation}
\end{ceqn}

where, $\phi(\mu_j,s_j)=\left[\sqrt{\mu_j^2+s_j^2}-\mu_j-s_j\right]$. We define the Jacobian of the KKT conditions of the original problem in a neighborhood of the local solution as:

\begin{ceqn}
    \begin{equation}
        J^{(k)}=\begin{bmatrix}
            \nabla^2_{yy}\mathcal{L}_\mathrm{orig} & J_h^\top & J_g^\top & 0\\
            J_h & 0 & 0 & 0\\
            J_g & 0 & 0 & I\\
            0 & 0 & D_\mu & D_s
        \end{bmatrix},
    \end{equation}
\end{ceqn}

where, $D_\mu=\mathrm{diag}(\mu_i/\sqrt{\mu_i^2+s_i^2}-1)$ and $D_s=\mathrm{diag}(s_i/\sqrt{\mu_i^2+s_i^2}-1)$. Due to the assumptions we made earlier, the Jacobian at iteration $k$, $J^{(k)}$ is non-singular and invertible. Similarly the KKT of the approximated problem (\ref{p_approx_conv}) is:

\begin{ceqn}
    \begin{equation}
        \bar{F}(z) = \begin{bmatrix}
            \nabla_y\mathcal{L}_\mathrm{approx}(y,\lambda,\mu)\\
            J_h(x,y^k)(y-y^k)+h(x,y^k)\\
            J_g(x,y^k)(y-y^k)+g(x,y^k)+s\\
            \phi(\mu,s)
        \end{bmatrix},
    \end{equation}
\end{ceqn}
the Jacobian of the KKT conditions of the approximated problem in a neighborhood of the local solution is: 

\begin{ceqn}
    \begin{equation}
        J^{(k)}=\begin{bmatrix}
            \nabla^2_{yy}\mathcal{L}_\mathrm{approx} & J_h^\top & J_g^\top & 0\\
            J_h & 0 & 0 & 0\\
            J_g & 0 & 0 & I\\
            0 & 0 & D_\mu & D_s
        \end{bmatrix},
    \end{equation}
\end{ceqn}

The deviation of the projection from complete SQP step can be expressed through a matrix $E$ where 

\begin{ceqn}
    \begin{equation}
        \begin{aligned}
            E^{(k)}=J^{(k)}-\bar{J}^{(k)}=\begin{bmatrix}
            \nabla^2_{yy}\mathcal{L}_\mathrm{orig} - \nabla^2_{yy}\mathcal{L}_\mathrm{approx} & 0 & 0 & 0\\
            0 & 0 & 0 & 0\\
            0 & 0 & 0 & 0\\
            0 & 0 & 0 & 0
        \end{bmatrix},
        \end{aligned}
    \end{equation}
\end{ceqn}

where, $\nabla^2_{yy}\mathcal{L}_\mathrm{orig} - \nabla^2_{yy}\mathcal{L}_\mathrm{approx} = \sum_i\lambda_i^{(k)}\nabla^2h_i(y^{(k)}) + \sum_j\mu_j^{(k)}\nabla^2g_j(y^{(k)})$, the deviation comes from the second order information of the constraints. At $k$th layer, we define the residual $r^{(k)}=F(z^{(k)})$, the error $e^{(k)}=z^{(k)}-z^*$ and solve for the Newton's step $\Delta z^{(k)}$:
\begin{ceqn}
    \begin{equation}
        \begin{aligned}
            \bar{J}\Delta z^{(k)}&=-r^{(k)}\\
            z^{(k+1)}&=z^{(k)}+\Delta z^{(k)}\\
            e^{(k+1)}&=e^{(k)}+\Delta z^{(k)}.
        \end{aligned}
    \end{equation}
\end{ceqn}

Using first order Taylor's expansion,

\begin{ceqn}
    \begin{equation}
        F(z^{(k)})= J^{(k)}e^{(k)}+r^{(k)},
    \end{equation}
\end{ceqn}

where, $r^{(k)}=\mathcal{O}(\lVert e^{(k)}\rVert)$. Therefore,

\begin{ceqn}
    \begin{equation}
        \bar{J}^{(k)}\Delta z^{(k)}=-r^{(k)}=-F(z^{(k)})=-J^{(k)}e^{(k)}-r^{(k)}.
    \end{equation}
\end{ceqn}
Since, $E^{(k)}=J^{(k)}-\bar{J}^{(k)}$,
\begin{ceqn}
    \begin{equation}
        \begin{aligned}
            \left(J^{(k)}-E^{(k)}\right)\Delta z^{(k)}&=-J^{(k)}e^{(k)}-r^{(k)},\\
            (J^{(k)})^{-1}\left(J^{(k)}-E^{(k)}\right)\Delta z^{(k)}&=-(J^{(k)})^{-1}\left(J^{(k)}e^{(k)}-r^{(k)}\right),\\
            (I-M)\Delta z^{(k)}&=-e^{(k)}-(J^{(k)})^{-1}r^{(k)},\\
            \Delta z^{(k)}&=-(I-M)^{-1}\left(e^{(k)}+(J^{(k)})^{-1}r^{(k)}\right),
        \end{aligned}
    \end{equation}
\end{ceqn}
where, $M=((J^{(k)})^{-1}E^{(k)})$. Hence, the error at ($k+1$)th layer can be expressed as:
\begin{ceqn}
    \begin{equation}
        \begin{aligned}
            e^{(k+1)}&=e^{(k)}+\Delta z^{(k)}\\
            &= e^{(k)}-(I-M)^{-1}\left(e^{(k)}+(J^{(k)})^{-1}r^{(k)}\right)\\
            &= \left(I-(I-M)^{-1}\right)e^{(k)}-(I-M)^{-1}(J^{(k)})^{-1}r^{(k)}\\
            &= \left((I-M)(I-M)^{-1}-(I-M)^{-1}\right)e^{(k)}-(I-M)^{-1}(J^{(k)})^{-1}r^{(k)}\\
            &= (I-M)^{-1}\left((I-M)-I\right)e^{(k)}-(I-M)^{-1}(J^{(k)})^{-1}r^{(k)}\\
            &= -(I-M)^{-1}Me^{(k)}-(I-M)^{-1}(J^{(k)})^{-1}r^{(k)}.
        \end{aligned}
    \end{equation}
\end{ceqn}

Using Banach's lemma ($\lVert(I-A)^{-1}\rVert\le 1/(1-\lVert A\rVert)$) i.e., $\lVert(I-M)^{-1}\rVert\le C_0$, and $r^{(k)}=\mathcal{O}(\lVert e^{(k)}\rVert)$ i.e., $\lVert r^{(k)}\rVert\le C_1\lVert e^{(k)}\rVert^2$ the error can be estimated,

\begin{ceqn}
    \begin{equation}\label{convergence_rate}
        \lVert e^{(k+1)}\rVert\le C_0(\lVert M\rVert)\lVert e^{(k)}\rVert + C_0\lVert (J^{(k)})^{-1}\rVert C_1\lVert e^{(k)}\rVert^2.
    \end{equation}
\end{ceqn}

From Eq. \ref{convergence_rate}, nearby $z^*$ i.e., $y^*$, when $M=0$, the projection yields quadratic convergence. It can be easily inferred this is true when the constraints are affine, only one layer is required to satisfy the original constraints. When $M\neq 0$ and $\lVert M\rVert<1$, linear convergence is guaranteed. However, when $\lVert M\rVert\geq1$, second order corrections might require to converge to the solution.
\section{Chambolle-Pock Algorithm}\label{stepbystepCP}
The Chambolle-Pock (CP) algorithm \cite{chambolle2011first} for solving the generic saddle-point problem of the following form

\begin{ceqn}
\begin{equation} \label{CP_problem} \tag{SPP}
\begin{aligned}
    \textit{Primal Problem:} & \quad \min_{y\in Y} F(Ky) + G(y)\\
    \textit{Dual Problem:} & \quad \max_{\lambda \in \Lambda} -(G^*(-K^*\lambda)+F^*(\lambda))
\end{aligned}
\end{equation}
\end{ceqn}

where, the map $K:Y\mapsto \Lambda$ is a continuous linear operator, $G: Y\mapsto[0,+\infty)$, $F^*:\Lambda\mapsto[0,+\infty)$ are proper, convex, lower semicontinuous functions while $F^*$ is the convex conjugate of the convex lower semicontinuous function $F$. The algorithm assumes the \ref{CP_problem} has at least a solution $(\hat{y},\hat{\lambda})\in Y\times \Lambda$, which satisfies

\begin{ceqn}
\begin{equation}
\begin{aligned}
    K\hat{y} &\in \partial F^*(\hat{\lambda})\\
    -(K^*\hat{\lambda}) &\in \partial G(\hat{y})
\end{aligned}
\end{equation}
\end{ceqn}

where $\partial F^*$ and $\partial G$ are the subgradient of $F^*$ and $G$. To solve the problem defined in Eq. The update of the CP algorithm for solving the \ref{CP_problem} is

\begin{ceqn}
\begin{equation}
    \begin{cases}
        \lambda^{n+1} = (I + \sigma \partial F^*)^{-1}(\lambda^n+\sigma K\bar{y}^n)\\
        y^{n+1} = (I + \tau \partial G)^{-1}(y^n - \tau K^*\lambda^{n+1})\\
        \bar{y}^{n+1} = y^{n+1} + \theta (y^{n+1} - y^n)
    \end{cases}
\end{equation}
\end{ceqn}
where, $\tau,\sigma>0$ and $\theta\in[0,1]$. The algorithm converges when $\theta=1$ and $\tau\sigma\lVert K\rVert^2\leq1$. See \cite{chambolle2011first} for details. We consider the QP of the following form:

\begin{ceqn}\label{DQP}
    \begin{equation}
        \begin{aligned}
            \min &\quad\frac{1}{2}y^\top Qy + c^\top y\\
            \mathrm{s.t.} & \quad Ay = b \\
            & \quad Cy \leq d \\
            & \quad l \leq y \leq u
        \end{aligned}
    \end{equation}
\end{ceqn}
where, $Q$ is a diagonal matrix such that $Q = \mathrm{diag}(q_1,q_2,\cdots, q_n)\in\mathbb{R}^{n\times n}$, $A\in \mathbb{R}^{n_\textrm{eq}\times n}$, $b\in \mathbb{R}^{n_\textrm{eq}}$, $C\in \mathbb{R}^{n_\textrm{ineq}\times n}$, $d\in \mathbb{R}^{n_\textrm{ineq}}$, $l\in(R\cup\{-\infty\})^n$, and $u\in(R\cup\{\infty\})^n$. 
By introducing the dual variables $\lambda\in\mathbb{R}^{n_\textrm{eq}}$, $\mu\in\mathbb{R}^{n_\textrm{ineq}}$ and $\alpha,\beta\in\mathbb{R}^n$ corresponding to the equality, inequality and bound constraints respectively the saddle-point problem becomes
\begin{ceqn}
    \begin{equation}\label{QP_Saddle}
        \min_{l\ \leq y\ \leq u} \quad  \max_{\lambda,\ \mu\geq 0} \quad \frac{1}{2}y^\top Qy + c^\top y + \lambda^\top(Ay-b) + \mu^\top(Cy-d). 
    \end{equation}
\end{ceqn}
For a general quadratic problem defined by Eq. \ref{QP_Saddle} the steps are generated as follows where the combined constraint matrix $K$, combined coefficient vector $h$ and combined multipliers $w$ are defined as $K^\top = (A^\top, C^\top)$, $h^\top = (b^\top, d^\top)$ and $w^\top=(\lambda^\top, \mu^\top)$. Multipliers for the bound constraints are obtained analytically.

\textbf{Step 1:} \textit{Dual update}\\
Updating the dual variable involves solving the convex quadratic problem (in $w$)

\begin{ceqn}
\begin{equation}
        w^{k+1} = \arg\min_{w} \frac{1}{2}\lVert w-(w^k+\sigma K \bar{y}^k)\rVert^2+\sigma h^\top w.
\end{equation}
\end{ceqn}

By setting the gradient to zero yields, $w - (w^k+\sigma K \bar{y}^k)+\sigma h = 0$. The update step becomes,

\begin{ceqn}
    \begin{equation}\label{dual_update_z}
        w^{k+1} = w^k + \sigma( K \bar{y}^k-h).
    \end{equation}
\end{ceqn}

To retain the complementary conditions ($\mu \geq 0$) in the dual variables, the update in Eq. \ref{dual_update_z} can be split into

\begin{ceqn}
    \begin{equation}\label{dual_update}
    \begin{aligned}
        \lambda^{k+1} &= \lambda^k + \sigma (A\bar{y}^k-b)\\
        \mu^{k+1} &= \max(0,\mu^k + \sigma (C\bar{y}^k-d))
    \end{aligned}
    \end{equation}
\end{ceqn}

\textbf{Step 2:} \textit{Primal Update}\\
Updating the primal variable involves solving the convex quadratic problem (in $y$)

\begin{ceqn}
    \begin{equation}
        \nu^{k+1} = \arg \min_{\nu} \frac{1}{2} \lVert \nu-(y^k-\tau K^\top w^{k+1})\rVert^2 + \tau\left(\frac{1}{2}\nu^\top Q \nu + c^\top \nu\right).
    \end{equation}
\end{ceqn}

By setting the gradient to zero yields, $\nu-(y^k-\tau K^\top w^{k+1})+\tau Q\nu+\tau c=0$ i.e. $(I+\tau Q)\nu = y^k - \tau K^\top z^{k+1}-\tau c$. The update step becomes,

\begin{ceqn}
    \begin{equation}\label{primal_update}
        \nu^{k+1} = P\left(y^k - \tau (A^\top\lambda^{k+1}+C^\top\mu^{k+1})-\tau c\right)
    \end{equation}
\end{ceqn}

where, $P = (I+\tau Q)^{-1}$. It is worthwhile to mention that this is true for any convex QP. However, if the $Q = \mathrm{diag}(q_1,q_2,\cdots,q_n) =\mathrm{diag}(q_i)$ then $(I+\tau Q)^{-1}=\mathrm{diag}(1/(1+\tau q_i))$. Hence, computational cost involved in computing $P$ becomes negligible which requires minimal binary operations only.\\

To retain the box constraint we apply the box projection $\Pi_{[l,u]}$ on $\nu$.

\begin{ceqn}
    \begin{equation}
        y^{k+1} = \Pi_{[l,u]}(\nu^{k+1})
    \end{equation}
\end{ceqn}

where, $\Pi_{[l,u]}(v) = \min(u,\max(l,v))$ is the projection to keep the primal feasibility on the box constraints.\\


\textbf{Step 3:} \textit{Update initialization}\\
\begin{ceqn}
    \begin{equation}
        \bar{y}^{k+1}=y^{k+1}+\theta(y^{k+1}-y^k)
    \end{equation}
\end{ceqn}
\textbf{Termination Criteria:} As \ref{subproblem} remains convex, the primal and dual feasibility and primal-dual gap fit perfectly as termination criteria to get the optimal solution of the layer $\mathcal{P}_s^i$. We set the termination criteria when the primal residuals, the dual residuals and the primal-dual gap are smaller than some predefined tolerance levels $\epsilon_{\mathrm{prim}}>0, \epsilon_{\mathrm{dual}}>0$, and $\epsilon_{\mathrm{gap}}>0$ respectively, i.e., 
\begin{ceqn}
    \begin{equation}
        \lVert r^k_{\mathrm{prim}}\rVert_\infty\leq\epsilon_{\mathrm{prim}}, \quad \lVert r^k_{\mathrm{dual}}\rVert_\infty\leq\epsilon_{\mathrm{dual}}, \quad \lVert r^k_{\mathrm{gap}}\rVert_\infty\leq\epsilon_{\mathrm{gap}},
    \end{equation}
\end{ceqn}
where,
\begin{ceqn}
    \begin{equation}
    \begin{aligned}
        r_{\mathrm{prim}}^k & = \begin{bmatrix}
                                A y^k - b \\
                                \max(C y^k - d,\; 0)
                                \end{bmatrix},\quad r_{\mathrm{dual}}^k = Qy^k+c+A^\top\lambda^k+C^\top\mu^k-\alpha^k+\beta^k,\\
        r_{\mathrm{gap}}^k & = (\lambda^k)^\top (A y^k - b) + (\mu^k)^\top (C y^k - d) + (\alpha^k)^\top (l - y^k) + (\beta^k)^\top (y^k - u).
    \end{aligned}
    \end{equation}
\end{ceqn}
We emphasize that the dual gap ($r^k_{\mathrm{dual}}$) computation may include errors due to an iterative projection on the box. The box projection multiplier values may not represent the actual multipliers until converged. Therefore during implementation, we only enforce primal feasibility and primal-dual gap. As the box projection is always feasible, the last two terms in the primal-dual gap ($(\alpha^k)^\top (l - y^k)$ and $ (\beta^k)^\top (y^k - u)$) are always zero. Taking the primal residual and primal-dual gap provides plausible criteria for termination.

\textbf{Convergence:} With the criteria on the algorithm parameters that for ($\theta=1$) and
\begin{ceqn}
    \begin{equation}
        \left(\frac{1}{\tau}-L_f\right)\frac{1}{\sigma}<\lVert K\rVert^2,
    \end{equation}
\end{ceqn}
where, $L_f$ is the Lipschitz constant (in our case the maximum value of the $Q$ matrix), $\lVert K\rVert$ is the spectral norm of $K$ ($K^\top = (A^\top,C^\top)$), the CP iterates converge to a primal-dual solution and the ergodic primal-dual gap decreases as $\mathcal{O}(1/k)$ where $k$ is the number of iteration \cite{chambolle2016ergodic}. The spectral norm of $K$ is computed with power iteration algorithm (provided in the Appendix \ref{power_iter}). The Algorithm \ref{CP_QP} can be further accelerated for strongly convex problems with less number of iterations by updating the parameters. We provide the accelerated method in the Appendix \ref{acc_CP}. However, due to change in parameters each iteration may require additional computation which limits the accelerated method. Additionally, we discuss the conditioning for the matrices in CP algorithm in Appendix \ref{conditioning}. In practice, conditioning reduces the number of iterations to converge the CP algorithm. Note that, when $Q\succeq0$, the CP algorithm converges to the optimal solution.

\textbf{Handling Non-Affine Constraints}: The CP algorithm is stated for more general case rather than only affine constraints. This leads us to have appropriate proximal operator (inequality multiplier update in Step 1) for convex constraint of specific form. For example, convex quadratic inequality constraints can be reformulated as second order conic constraint and can be solved exactly in the CP algorithm with a few change. We provide such a modified version of Algorithm \ref{CP_QP} for quadratic inequality constraint which has exploitable convex structure (see Appendix \ref{qcqp_reform}). If the nonlinear constraints have exploitable structure then similar modification can be made in the proximal operator for dual update. The advantage of this is that it satisfies the nonlinear constraints in single subproblem i.e., $k=1$, so that the computational burden reduces from solving multiple subproblems to one subproblem.

\section{Accelerated Chambolle-Pock algorithm}\label{acc_CP}
When, $G$ is uniformly convex in \ref{CP_problem}, the rate of convergence of the Chambolle-Pock algorithm can be improved to $\mathcal{O}(1/k^2)$, by choosing the parameters of the algorithm iteratively instead of a fixed value \cite{chambolle2016ergodic}. Algorithm \ref{alg:CP_acc} is the accelerated version of the CP algorithm. Here, $\gamma$ is the uniform convexity constant, for QP which is the $\lambda_\mathrm{min}(Q)=\mathrm{min}(Q)$ as $Q$ is diagonal.

\begin{algorithm}[h]
\small
\caption{Accelerated Chambolle-Pock for QPs}\label{alg:CP_acc}
\DontPrintSemicolon
\SetKwBlock{CPQP}{\texttt{AcceleratedChambollePock}:}{}

\CPQP{
\KwIn{Initial values $y^0,\lambda^0,\mu^0 $, $p$}
\Problem{$\min \tfrac{1}{2}y^\top Qy+c^\top y;\quad \mathrm{s.t.}\ Ay=b; \ Cy \leq d; \ l\leq y\leq u$ }
\Extract{$Q, c, A, b, C, d, l ,u \leftarrow p$}
\Parameter{$\tau>0, \sigma>0, \theta=1$ while $\left(\frac{1}{\tau}-L_f\right)\frac{1}{\sigma}<\lVert K\rVert^2$}
\Compute{$P=\mathrm{diag}(1/(1+\tau q_i)), \gamma = \max\!\big(\min_i q_i,0\big)$}\;
\Initialize{$k\leftarrow0, y^0\leftarrow y^0,\lambda^0\leftarrow \lambda^0,\mu^0\leftarrow \mu^0,\bar{y}^0\leftarrow y^0$}\;

\Repeat{termination criteria is satisfied}{

        $\lambda^{k+1} = \lambda^k + \sigma (A\bar{y}^k-b)$\;
        
        $\mu^{k+1} = \max(0,\mu^k + \sigma (C\bar{y}^k-d))$\;

        $\nu^{k+1} = P\big(y^k - \tau (A^\top\lambda^{k+1}+C^\top\mu^{k+1})-\tau c\big)$\;

        $y^{k+1} = \Pi_{[l,u]}(\nu^{k+1})$


        $\bar{y}^{k+1}=y^{k+1}+\theta(y^{k+1}-y^k)$\;
        
        \If{$\gamma>0$}{
        
        $\theta \leftarrow \dfrac{1}{\sqrt{1+\gamma\tau}}$\;
        
        $\tau \leftarrow \theta\tau$\;
        
        $\sigma \leftarrow \dfrac{\sigma}{\theta}$\;
        
        $P \leftarrow \mathrm{diag}\!\left(\dfrac{1}{1+\tau q_i}\right)$\;
        }

        $k\leftarrow k+1$
        }
\KwOut{$z\leftarrow y^{k},\lambda^k,\mu^k$}
}
\end{algorithm}

\section{Power iteration for spectral norm computation}\label{power_iter}
To estimate the operator norm required for step-size selection, we employ a randomized power iteration procedure on the matrix $K^\top K$, where
\begin{ceqn}
    \begin{equation}
        K=\begin{bmatrix}A\\C\end{bmatrix}.
    \end{equation}
\end{ceqn}
The method starts from a randomly initialized normalized vector and repeatedly applies the map $x \mapsto K^\top Kx = A^\top Ax + C^\top Cx$. At each iteration, the vector is normalized, and the corresponding Rayleigh quotient is used to approximate the dominant eigenvalue of $K^\top K$. The spectral norm is then estimated as $\|K\| \approx \sqrt{\lambda_{\max}(K^\top K)}$. This avoids explicitly forming $K^\top K$ and provides an efficient procedure for estimating the quantity needed in the primal-dual step size condition. Algorithm \ref{alg:power_iteration} algorithmic steps to estimate the norm.

\begin{algorithm}[h]
\small
\caption{Power Iteration for Estimating $\|K\|$}
\label{alg:power_iteration}
\DontPrintSemicolon
\SetKwBlock{PI}{\texttt{EstimateKNorm}:}{}

\PI{
\KwIn{$A,\ C$, random key, number of iterations $N$}
\Parameter{$\varepsilon>0$}
\Initialize{$\mathrm{Sample\ a\ random\ vector\ } x^0$}\;
\Initialize{$k\leftarrow 0$, $x^0 \leftarrow \frac{x^0}{\|x^0\|+\varepsilon}$}

\Repeat{$k = N$}{
    $a^k \leftarrow Ax^k$\;
    $c^k \leftarrow Cx^k$\;
    $y^k \leftarrow A^\top a^k + C^\top c^k$\;
    $\lambda^k \leftarrow (x^k)^\top y^k$\;
    $x^{k+1} \leftarrow \dfrac{y^k}{\|y^k\|+\varepsilon}$\;
    $k \leftarrow k+1$\;
}
\KwOut{$\|K\| \leftarrow \sqrt{\max(\lambda^{k-1},0)}+\varepsilon$}
}
\end{algorithm}
\section{Conditioning of the matrices in Chambolle-Pock}\label{conditioning}
Preconditioning such as Ruiz's equilibration\cite{ruiz2001scaling} is often used as a heuristic for primal-dual algorithms\cite{stellato2020osqp}. Ruiz equilibration is applied as a preprocessing step to improve the numerical conditioning of the quadratic program before running the Chambolle--Pock iterations. The idea is to iteratively rescale the columns and rows of the problem data so that their magnitudes become more balanced. In the present setting, column scaling is applied to the primal variables, while separate row scalings are applied to the equality and inequality constraint matrices. This produces a scaled problem with better-balanced coefficients, which typically reduces the number of iterations in Chambolle-Pock algorithm. After solving the scaled problem, the primal and dual variables are mapped back to the original coordinates using the accumulated scaling factors.

Let $S\in\mathbb{R}^{n\times n}$ denote the diagonal column-scaling matrix for the primal variables, and let $R_\mathrm{eq}\in\mathbb{R}^{n_\mathrm{eq}\times n_\mathrm{eq}}$ and $R_\mathrm{ineq}\in\mathbb{R}^{n_\mathrm{ineq}\times n_\mathrm{ineq}}$ denote the diagonal row-scaling matrices for the equality and inequality constraints, respectively. The scaled quadratic program is written in terms of the transformed primal variable $y_s$, defined through $y = Sy_s$.
Under this transformation, the scaled problem data become:
\begin{ceqn}
    \begin{equation}
        Q_s = SQS,\quad c_s = Sc,\quad A_s = R_{\mathrm{eq}}AS,\quad b_s = R_{\mathrm{eq}}b,\quad C_s = R_{\mathrm{ineq}}CS,\quad d_s = R_{\mathrm{ineq}}d,        
    \end{equation}
\end{ceqn}
with the bound vectors transformed as
\begin{ceqn}
    \begin{equation}
        l_s = S^{-1}l,\qquad u_s = S^{-1}u.
    \end{equation}
\end{ceqn}
The Chambolle-Pock iterations are then performed on the scaled problem to obtain a solution $(y_s^*,\lambda_s^*,\mu_s^*)$. After convergence, the solution is mapped back to the original coordinates by reversing the applied scalings. In particular, the primal variable is recovered as $y^* = Sy_s^*$, while the dual variables associated with the equality and inequality constraints are recovered as $\lambda^* = R_\mathrm{eq}\lambda_s^*$ and $\mu^* = R_\mathrm{ineq}\mu_s^*$. Thus, the optimization is carried out entirely in the equilibrated space, but the final primal and dual solutions are returned in the original problem coordinates. Algorithm \ref{alg:ruiz_equilibration} provides the step by step procedure we used for equilibration.

\begin{algorithm}[h]
\small
\caption{Modified Ruiz Equilibration for Box-Constrained QPs}
\label{alg:ruiz_equilibration}
\DontPrintSemicolon
\SetKwBlock{RUIZ}{\texttt{RuizEquilibrate}:}{}

\RUIZ{
\KwIn{$Q,c,A,b,C,d,l,u$, number of iterations $N$, tolerance $\varepsilon>0$}
\Problem{$\min \ \tfrac{1}{2}y^\top Qy + c^\top y \quad \mathrm{s.t.}\quad Ay=b,\ Cy\le d,\ l\le y\le u$}
\Compute{$n\leftarrow \mathrm{length}(Q),\ n_\mathrm{eq}\leftarrow \mathrm{rows}(A),\ n_\mathrm{ineq}\leftarrow \mathrm{rows}(C)$}\;
\Initialize{$Q_s \leftarrow Q,\ c_s \leftarrow c,\ A_s \leftarrow A,\ b_s \leftarrow b,\ C_s \leftarrow C,\ d_s \leftarrow d,\ l_s \leftarrow l,\ u_s \leftarrow u$\\
\hspace*{4.5em}$S \leftarrow I_n,\ R_e \leftarrow I_{n_\mathrm{eq}},\ R_i \leftarrow I_{n_\mathrm{ineq}}$\\
\hspace*{4.5em}$k \leftarrow 0$}

\Repeat{$k = N$}{
    \tcp{Column equilibration}
    \Compute{$(m_{\mathrm{col}})_j = \max\!\Big(\max_i |(A_s)_{ij}|,\ \max_i |(C_s)_{ij}|,\ |(Q_s)_j|\Big),\ \forall j$}\;
    \Compute{$\delta_j = \dfrac{1}{\sqrt{\max((m_{\mathrm{col}})_j,\varepsilon)}},\ \forall j$}\;

    $A_s \leftarrow A_s\,\mathrm{diag}(\delta), C_s \leftarrow C_s\,\mathrm{diag}(\delta), Q_s \leftarrow Q_s \odot \delta^2, c_s \leftarrow c_s \odot \delta, l_s \leftarrow l_s \oslash \delta, u_s \leftarrow u_s \oslash \delta$\;
    $S \leftarrow S\,\mathrm{diag}(\delta)$\;

    \tcp{Equality-row equilibration}
    \If{$n_\mathrm{eq} \neq 0$}{
        \Compute{$(m_{\mathrm{eq}})_i = \max_j |(A_s)_{ij}|,\ \forall i$}\;
        \Compute{$\eta_i = \dfrac{1}{\sqrt{\max((m_{\mathrm{eq}})_i,\varepsilon)}},\ \forall i$}\;
        $A_s \leftarrow \mathrm{diag}(\eta)\,A_s, b_s \leftarrow \eta \odot b_s$\;
        $R_e \leftarrow \mathrm{diag}(\eta)\,R_e$\;
    }

    \tcp{Inequality-row equilibration}
    \If{$n_\mathrm{ineq} \neq 0$}{
        \Compute{$(m_{\mathrm{ineq}})_i = \max_j |(C_s)_{ij}|,\ \forall i$}\;
        \Compute{$\zeta_i = \dfrac{1}{\sqrt{\max((m_{\mathrm{ineq}})_i,\varepsilon)}},\ \forall i$}\;
        $C_s \leftarrow \mathrm{diag}(\zeta)\,C_s, d_s \leftarrow \zeta \odot d_s$\;
        $R_i \leftarrow \mathrm{diag}(\zeta)\,R_i$\;
    }

    $k \leftarrow k+1$\;
}
\KwOut{$Q_s,c_s,A_s,b_s,C_s,d_s,l_s,u_s,S,R_e,R_i$}
}
\end{algorithm}
\section{Reformulation for Convex Quadratic Constraints}\label{qcqp_reform}
The projection layer requires solving $k$ subproblems to get feasible solution. However, solving a single subproblem is sufficient for affine constraints as well as specific classes of nonlinear constraints, such as second order conic constraints because of having appropriate proximal operator in the solving step. Quadratic constraints can be reformulated as second order conic constraints. For each quadratic inequality
\begin{ceqn}
    \begin{equation}
    y^\top C_i y + d_i^\top y \le e_i - E_i x,
    \label{eq:quad_constraint_original}
    \end{equation}    
\end{ceqn}

let $C_i = R_i^\top R_i$ be a factorization of \(C_i\). Assume \(d_i \in \mathrm{Range}(C_i)\), and define  $q_i := \frac{1}{2}(R_i^\dagger)^\top d_i$, where \(R_i^\dagger\) denotes the Moore-Penrose pseudo-inverse. Then
\begin{ceqn}
    \begin{equation}
    y^\top C_i y + d_i^\top y = \|R_i y + q_i\|_2^2 - \|q_i\|_2^2.
    \label{eq:complete_square}
    \end{equation}
\end{ceqn}

Substituting \eqref{eq:complete_square} into \eqref{eq:quad_constraint_original} yields $\|R_i y + q_i\|_2^2 \le e_i - E_i x + \|q_i\|_2^2$. Defining the radius $r_i(x) := \sqrt{e_i - E_i x + \|q_i\|_2^2}$ assuming $r_i(x)^2 \ge 0$. Then the quadratic inequality becomes
\begin{ceqn}
    \begin{equation}
    \|R_i y + q_i\|_2 \le r_i(x), \qquad i=1,\dots,m.
    \label{eq:ball_constraint}
    \end{equation}    
\end{ceqn}

Lets consider the parametric convex QCQP
\begin{ceqn}
    \begin{equation}\label{eq:qcqp_original}
    \begin{aligned}
    \min_{y}\quad & \frac{1}{2}y^\top Qy + c^\top y \\
    \mathrm{s.t.}\quad & Ay = b + Bx, \\
    & y^\top C_i y + d_i^\top y \le e_i - E_i x, \qquad i=1,\dots,m, \\
    & l + Lx \le y \le u + Ux,
    \end{aligned}
    \end{equation}
\end{ceqn}

where \(Q \succeq 0\) and \(C_i \succeq 0\) for all \(i=1,\dots,m\). With the reformulation the problem in Eq. \ref{eq:qcqp_original} can be rewritten as

\begin{ceqn}
    \begin{equation}
    \begin{aligned}
    \min_{y}\quad & \frac{1}{2}y^\top Qy + c^\top y \\
    \mathrm{s.t.}\quad
    & Ay = b + Bx, \\
    & \|R_i y + q_i\|_2 \le r_i(x), \qquad i=1,\dots,m, \\
    & l + Lx \le y \le u + Ux.
    \end{aligned}
    \label{eq:qcqp_reformulated}
    \end{equation}
    \end{ceqn}

\textbf{Primal-Dual Saddle Form}\\

The indicator of the Euclidean ball \(\mathbb{B}(r_i(x)) := \{z : \|z\|_2 \le r_i(x)\}\) has conjugate
\begin{ceqn}
    \begin{equation}
I_{\mathbb{B}(r_i(x))}^*(\mu_i) = r_i(x)\|\mu_i\|_2.
\end{equation}
\end{ceqn}

Using this identity, the reformulated problem \eqref{eq:qcqp_reformulated} admits the saddle-point representation

\begin{ceqn}
    \begin{equation}
    \begin{aligned}
    \min_{y \in [\,l+Lx,\;u+Ux\,]}\max_{\lambda,\mu_i}
    \quad &
    \frac{1}{2}y^\top Qy + c^\top y
    + \lambda^\top\!\big(Ay-(b+Bx)\big) \\
    &\qquad
    + \sum_{i=1}^m \mu_i^\top (R_i y + q_i)
    - \sum_{i=1}^m r_i(x)\|\mu_i\|_2 .
    \end{aligned}
    \label{eq:saddle_form}
    \end{equation}
\end{ceqn}

The corresponding primal proximal map is $\operatorname{prox}_{\tau G}(v) = (I+\tau Q)^{-1}(v-\tau c)$, where $G(y) := \frac{1}{2}y^\top Qy + c^\top y$. If \(Q\) is diagonal with diagonal entries \(q_j\), then the proximal map reduces to
\begin{ceqn}
    \begin{equation}
    P := \operatorname{diag}\!\left(\frac{1}{1+\tau q_j}\right),
    \qquad
    \operatorname{prox}_{\tau G}(v)=P(v-\tau c).
    \end{equation}
\end{ceqn}
    
For each dual block, the proximal operator of \(\sigma r_i(x)\|\cdot\|_2\) is the vector soft-thresholding map
\begin{ceqn}
    \begin{equation}
    \operatorname{prox}_{\sigma r_i(x)\|\cdot\|_2}(v) =
    \begin{cases}
    \left(1-\dfrac{\sigma r_i(x)}{\|v\|_2}\right)v, & \|v\|_2 > \sigma r_i(x), \\[2mm]
    0, & \|v\|_2 \le \sigma r_i(x).
    \end{cases}
    \label{eq:vector_shrinkage}
    \end{equation}
\end{ceqn}

The Chambolle-Pock algorithm for \eqref{eq:qcqp_reformulated} is given in Algorithm \ref{CP_QCQP}.\\

\begin{algorithm}[H]
\small
\caption{Chambolle-Pock for Convex QCQPs}\label{CP_QCQP}
\DontPrintSemicolon
\KwIn{Initial values $y^0,\lambda^0,\mu_1^0,\dots,\mu_m^0$}
\Problem{$\min \tfrac{1}{2}y^\top Qy+c^\top y;\quad \mathrm{s.t.}\ Ay=b+Bx;\ \|R_i y+q_i\|_2\le r_i(x),\ i=1,\dots,m;\ l+Lx\le y\le u+Ux$}
\Parameter{$\tau>0,\sigma>0,\theta\in[0,1]$}
\Compute{$P=(I+\tau Q)^{-1}$ \tcp*{or $P=\mathrm{diag}(1/(1+\tau q_j))$ if $Q$ is diagonal}}
\Initialize{$k\leftarrow0,\ y^0\leftarrow y^0,\ \lambda^0\leftarrow \lambda^0,\ \mu_i^0\leftarrow \mu_i^0\ (i=1,\dots,m),\ \bar y^0\leftarrow y^0$}

\Repeat{termination criteria is satisfied}{

        $\lambda^{k+1} = \lambda^k + \sigma \big(A\bar y^k-(b+Bx)\big)$\;
        
        \For{$i=1,\dots,m$}{
            $v_i^k = \mu_i^k + \sigma (R_i\bar y^k+q_i)$\;
            
            $\mu_i^{k+1} =
            \begin{cases}
            \left(1-\dfrac{\sigma r_i(x)}{\|v_i^k\|_2}\right)v_i^k, & \|v_i^k\|_2 > \sigma r_i(x), \\[2mm]
            0, & \|v_i^k\|_2 \le \sigma r_i(x)
            \end{cases}$\;
        }

        $g^{k+1} = A^\top\lambda^{k+1} + \sum_{i=1}^m R_i^\top \mu_i^{k+1}$\;

        $y^{k+1} = \Pi_{[\,l+Lx,\;u+Ux\,]}\big(P(y^k - \tau g^{k+1} - \tau c)\big)$\;

        $\bar{y}^{k+1}=y^{k+1}+\theta(y^{k+1}-y^k)$\;

        $k\leftarrow k+1$
        }
\KwOut{$y^{k},\lambda^k,\mu_1^k,\dots,\mu_m^k$}
\end{algorithm}

\textbf{Remarks on the Algorithmic Structure:} Compared with the Chambolle-Pock scheme for linearly constrained QPs, the equality update preserves the same form, except that the right-hand side becomes parameter dependent through \(b+Bx\). The box projection also preserves the same structure, with shifted bounds \(l+Lx\) and \(u+Ux\). The only essential modification occurs in the inequality dual step: the linear inequality update based on the componentwise positive-part operator is replaced by a blockwise proximal update \eqref{eq:vector_shrinkage}, which arises from the conjugate of the Euclidean ball indicator after completing the square in each quadratic constraint.

If desired, the projection onto the parameterized box can be written componentwise as
\begin{ceqn}
    \begin{equation}
\Pi_{[\,l+Lx,\;u+Ux\,]}(v)
=
\min\!\big(\max(v,\;l+Lx),\;u+Ux\big).
\end{equation}
\end{ceqn}

Although it is possible to incorporate different constraint directly in the CP algorithm, in our implementation we keep linearized constraints for generalization of the framework.
\section{Primal-Dual solution of approximated problem}\label{linearization}
Let, the convex (nonlinear) optimization problem is defined as
\begin{ceqn}
\begin{equation} 
\label{eq:nlpconv} 
\begin{aligned} 
\min_{y \in \mathbb{R}^n} \quad & f(y) \\ 
\mathrm{s.t.}\quad & h(y) = 0, \\ 
& g(y) \le 0, 
\end{aligned}
\tag{P1}
\end{equation}
\end{ceqn}
where, $f$ is convex $\mathcal{C}^2$, $g$ is convex $\mathcal{C}^1$ and $h$ is affine. We assume that Slater's condition holds and $y^*$ is the global minimizer of this problem. With first order Taylor approximation of the constraints and the second order Taylor approximation of the objective at a arbitrary point $y_k$ we form the subproblem:
\begin{ceqn}
\begin{equation} 
\label{eq:nlp_sub} 
\begin{aligned} 
\min_{y \in \mathbb{R}^n} \quad & f(y_k) + \nabla f(y_k)^\top(y-y_k)+\frac{1}{2}(y-y_k)^\top H_d(y_k)(y-y_k) \\ 
\mathrm{s.t.} \quad & h(y) = 0, \\ 
& g(y_k)+\nabla g(y_k)^\top(y-y_k) \le 0,
\end{aligned}
\tag{P2}
\end{equation}    
\end{ceqn}

where, instead of the exact Hessian of the objective, we take only the diagonal hessian elements multiplied by a positive constant i.e., $H_d(y_k) = \rho \times \mathrm{diag}(\nabla^2f(y_k))$ with $\rho>0$.

\textbf{Lemma \ref{linearization}.1:} Second order Taylor approximation of a convex function is also convex.

\textbf{Proof:} If $f:\mathbb{R}^n\mapsto\mathbb{R}$ is convex then $\nabla^2f(y)\succcurlyeq0$ for all $y$. Second order Taylor approximation at $y_0$ of $f$ is defined as
\begin{ceqn}
\begin{equation}
    T_2(y) = f(y_0) + \nabla f(y_0)^\top(y-y_0)+\frac{1}{2}(y-y_0)^\top\nabla^2f(y_0)(y-y_0)
\end{equation}
\end{ceqn}

The hessian of $T_2(y)$ is $\nabla^2T_2(y) = \nabla^2f(y_0)$. Since $f$ is convex, $\nabla^2f(y_0)\succcurlyeq0$. Therefore $T_2(y)$ is convex. \hfill $\Box$

\textbf{Proposition:} The quadratic approximation of the second order Taylor approximation of a convex function with only the diagonal Hessian (The quadratic approximation of the objective in \ref{eq:nlp_sub}) is also convex.

\textbf{Proof:} As the hessian of the second order Taylor approximation is positive semi-definite (PSD), the diagonal hessian is also PSD, therefore convex.

\textbf{Lemma 2:} If $y^*$ is a global solution of the convex problem \ref{eq:nlpconv} and $\lambda^*, \mu^*$ are the dual solution, the subproblem \ref{eq:nlp_sub} has the same primal-dual solution when $y_k=y^*$.

\textbf{Proof:} According to lemma 1, the subproblem is also convex. Therefore if a any point satisfies KKT conditions then it would be sufficient for global optimality. At $y_k=y^*$ Lagrangian for the original and the subproblem becomes

\begin{equation}
\begin{aligned}
    \mathcal{L}_{P1} = &f(y) + \lambda^\top \nabla h(y) + \mu^\top g(y)\\
    \mathcal{L}_{P2} = &f(y^*) + \nabla f(y^*)^\top(y-y^*)+\frac{1}{2}(y-y^*)^\top H_d(y^*)(y-y^*) + \lambda^\top h(y)\\
    &+ \mu^\top\left[g(y^*)+\nabla g(y^*)^\top(y-y^*)\right]
\end{aligned}
\end{equation}

Assuming $x^*$ is the global minimizer of \ref{eq:nlpconv} and there exists unique dual variables any $\lambda^*$ and  $\mu^*\ge0$. Then the KKT conditions for \ref{eq:nlpconv} becomes

\begin{equation}
\label{nlpkkt}
\begin{aligned}
    \textit{Stationarity:} \quad \quad & \nabla f(y^*) + \sum_{i=1}^{n_\mathrm{eq}}\lambda_i^* \nabla h_i(y^*) + \sum_{j=1}^{n_\mathrm{ineq}}\mu_j^* \nabla g_j(y^*) = 0\\
    \textit{Primal feasibility:} \quad \quad & g(y^*)\leq 0,\quad  h(y^*)=0\\
    \textit{Dual feasibility:} \quad \quad & \mu\ge0\\
    \textit{Complementary slackness:} \quad \quad & \mu^* \odot g(y^*) = 0.
\end{aligned}
\end{equation}

Defining $g(y^*)+\nabla g(y^*)^\top(y-y^*) = \tilde{g}(y)$ the KKT conditions for the subproblem becomes

\begin{equation}
\label{subkkt}
\begin{aligned}
    \textit{Stationarity:} \quad \quad & \nabla f(y^*) + H_d(y^*)(y-y^*) + \sum_{i=1}^{n_\mathrm{eq}}\lambda_i \nabla h_i(y) + \sum_{j=1}^{n_\mathrm{ineq}}\mu_j^* \nabla g_j(y^*) = 0\\
    \textit{Primal feasibility:} \quad \quad & \tilde{g}(y)\leq 0, h(y)=0\\
    \textit{Dual feasibility:} \quad \quad & \mu\ge0\\
    \textit{Complementary slackness:} \quad \quad & \mu \odot \tilde{g}(y) = 0.
\end{aligned}
\end{equation}

At $y=y^*$, $\tilde{g}(y^*)=g(y^*)+\nabla g(y^*)^\top(y^*-y^*)=g(y^*)\le0$, $h(y^*)=0$ as $y^*$ is a feasible and optimal solution of original problem. Putting $y=y^*$ in the KKT system (Eq. \ref{subkkt}) satisfies the primal feasibility and yields a linear system for the dual variables. By inspection, $\lambda^*,\mu^*$ is the solution for the system. Therefore ($y^*,\lambda^*,\mu^*$) is a KKT point of the subproblem. As the problem is convex it is sufficient that $y^*$ is also a global minimizer for \ref{eq:nlp_sub}. Moreover, both the problems \ref{eq:nlpconv} and \ref{eq:nlp_sub} attains the same dual variables $\lambda^*$ and $\mu^*$. \hfill $\Box$

The motivation behind the lemma is to show that when the projection is done at near optimal solution (when the NN predicts close neighborhood), the projection pushes for optimality with obtaining primal-dual variables.

\begin{figure}[H]
    \centering
    \includegraphics[width=1\linewidth]{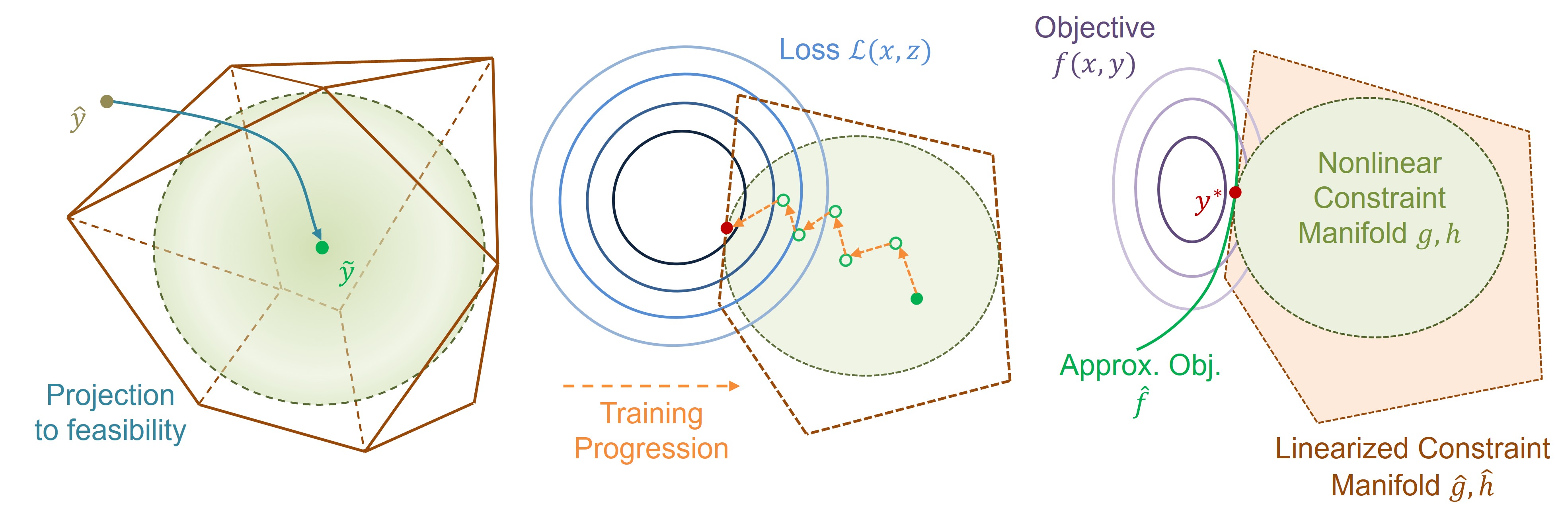}
    \caption{Graphical interpretation of the approximated layer. Left: the linearized model represents a polytope confining the original convex feasible domain. Middle: The neural network is learning to minimize the loss within the polytope, Right: the local exactness of the problem after linearization.}
    \label{fig:graphical}
\end{figure}

\textbf{Graphical interpretation:} We illustrate the complex events happening in the \texttt{NLPOpt-Net} framework graphically in Fig \ref{fig:graphical}. The illustrations represent the following phenomena: (left) relaxation of the original feasible domain: linearization creates a polytope defining the feasible region; (middle) the neural network is trained to minimize the loss within the polytope. The training pushes towards finding a better solution inside the polytope. Due to local convergence after $k$-layer, the projected output will be on the facet of the polytope. Moreover, due to the descent property of the projection layer, the solution after projection will remain no worse than raw backbone prediction (in practice better in most cases); (right) an $\epsilon$-optimal solution of the original problem \ref{Problem} can be obtained if the approximation is done near optimal point.

\section{Backward gradient of the projection}\label{backwardGrad}
\textbf{Projection as a Two Stage Mapping:} The projection layer involves two tasks in sequence. The first is to create the matrices $Q, A, C$ and the vectors $c, b, d, l, u$ which define the QPs in \ref{subproblem}. These matrices and vectors depend on the input x and network prediction $\hat{z}$. The second task is to solve the QPs using the CP algorithm. As the QPs in \ref{subproblem} are convex, they converge with arbitrary initial points i.e., the solution $\tilde{z}$ does not directly depend on the initialization $\hat{z}$ but on $\mathcal{M}(x,z)$ where $\mathcal{M}(x,z)\equiv(Q,c,A,b,C,d,l,u)$. The operator $\mathcal{M}$ defines the conversion of the subproblem. The differentiability still holds due to the composition of the projection. Since $\mathcal{P}_o(x,z) = \mathcal{P}_o(\mathcal{M}(x,z))$, chain rule leads to
\begin{ceqn}
    \begin{equation}\label{diff}
        \frac{\partial \mathcal{P}_o(x,z)}{\partial z}\Bigg|_{z=\hat{z}} = \frac{\partial \mathcal{P}_o(w)}{\partial w}\Bigg|_{w = \mathcal{M}(x,\hat{z})}\ \frac{\partial \mathcal{M}(x,z)}{\partial z}\Bigg|_{z=\hat{z}}.
    \end{equation}
\end{ceqn}
The second term of Eq. \ref{diff} can be obtained through automatic differentiation, which is not computationally expensive. We, therefore, provide the route for custom backward gradient computation of the first term.

\textbf{Fixed-Point Formulation:} Consider a fixed-point iteration of the form:
\begin{ceqn}
    \begin{equation}
        z^{k+1} = F(z^k;w),
    \end{equation}
\end{ceqn}
where, $z\in\mathbb{R}^m$ denotes the vector of all primal and dual variables,  $w=\mathcal{M}(x,z)\equiv(Q,c,A,b,C,d,l,u)$ denotes the collection of matrices and vectors defining the problem, and $F$ is a continuously differentiable neighborhood of a fixed point. When the iteration converges to $z^*$, it satisfies $z^* = F(z^*;w)$. Defining the residual mapping $G$ such that
\begin{ceqn}
    \begin{equation}\label{GF}
        G(z;w):=F(z;w)-z,        
    \end{equation}
\end{ceqn}
the fixed point satisfies $G(z^*;w)=0$. The projection output can be then extracted as $\tilde{z}$, i.e., $\tilde{z} = [y^{*\top}, \lambda^{*\top},\mu^{*\top}]^\top$.

\textbf{Deploying Implicit Function Theorem:} Assuming $F$ is continuously differentiable and the Jacobian is nonsingular, the implicit function theorem gives,
\begin{ceqn}
    \begin{equation}
        \frac{\partial G(z,w)}{\partial z}\ \frac{\partial z}{\partial w}\Bigg|_{z=z^*} + \frac{\partial G (z, w)}{\partial w}\Bigg|_{z=z^*} \implies \frac{\partial z^*}{\partial w} = -\left(\frac{\partial G(z,w)}{\partial z}\Bigg|_{z=z^*}\right)^{-1}\frac{\partial G (z, w)}{\partial w}\Bigg|_{z=z^*}.
    \end{equation}
\end{ceqn}
Since $\frac{\partial G}{\partial z} = \frac{\partial F}{\partial z}-I$ (from Eq. \ref{GF}), we obtain
\begin{ceqn}
    \begin{equation}
        \frac{\partial z^*}{\partial w} = \big(I - J_F(z^*)\big)^{-1}\frac{\partial F}{\partial w},
    \end{equation}
\end{ceqn}
where, $J_F(z^*) = \frac{\partial F}{\partial z}\Big|_{z=z^*}$. Now considering the loss function $\mathcal{L}(x,z^*)$ depend on the fixed point we get,
\begin{ceqn}
    \begin{equation}
        \frac{\partial \mathcal{L}}{\partial w}=g_z\frac{\partial z^*}{\partial w}
    \end{equation}
\end{ceqn}
where, $g_z = \frac{\partial \mathcal{L}}{\partial z}\Big|_{z=\tilde{z}}$. Hence, the implicit expression gives,
\begin{ceqn}
    \begin{equation}\label{ift1}
        \frac{\partial \mathcal{L}}{\partial w} = g_z^\top \big(I - J_F(z^*)\big)^{-1}\frac{\partial F}{\partial w}
    \end{equation}
\end{ceqn}
\textbf{Adjoint Formulation (VJP):} The inversion may not be plausible in Eq. \ref{ift1}. Even if exists, in practice, constructing and computing the inverse may be expensive for high dimensions. Instead of inverting, we solve the following adjoint linear system:
\begin{ceqn}
    \begin{equation}
        \left(I-J_F(z^*)^\top\right)v=g_z.
    \end{equation}
\end{ceqn}
The gradient with respect to $p$ is then given by
\begin{ceqn}
    \begin{equation}
        \frac{\partial \mathcal{L}}{\partial w}\Bigg|_{w=\mathcal{M}(x,\hat{z})}=v^\top\frac{\partial F}{\partial w}.
    \end{equation}
\end{ceqn}
We use the JAX \cite{jax2018github} implementation of bi-conjugate gradient stable iteration to get the vector $v$.


\end{document}